\documentclass{article}

\usepackage[final,nonatbib]{neurips_2022}

\usepackage{amsmath,bm}
\usepackage[utf8]{inputenc} 
\usepackage[T1]{fontenc}    
\usepackage{hyperref}       
\usepackage{url}            
\usepackage{booktabs}       
\usepackage{amsfonts}       
\usepackage{nicefrac}       
\usepackage{microtype}      
\usepackage{xcolor}   
\usepackage{multirow}
\usepackage{multicol}
\usepackage{wrapfig}
\usepackage{algorithmic}
\usepackage{algorithm}
\usepackage{graphics}
\usepackage{graphicx}

\newtheorem{proposition}{Proposition}
\newtheorem{definition}{Definition}
\newtheorem{remark}{Remark}
\newcommand{\maxSW}{\text{Max-SW} }
\newcommand{\ASW}{\mathcal{A}\text{-SW} }
\newcommand{\LASW}{\mathcal{LA}\text{-SW} }
\newcommand{\GASW}{\mathcal{GA}\text{-SW} }
\newcommand{\NASW}{\mathcal{NA}\text{-SW} }

\newcommand{\sphere}[1]{\mathbb{S}^{#1-1}}

\def\argmin{\textnormal{arg} \min}

\newcommand{\widgraph}[2]{\includegraphics[keepaspectratio,width=#1]{#2}}

\newcommand{\notiff}{%
  \mathrel{{\ooalign{\hidewidth$\not\phantom{"}$\hidewidth\cr$\iff$}}}}

\DeclareMathOperator*{\argmax}{arg\,max}

\title{Amortized Projection Optimization for Sliced Wasserstein Generative Models}

\author{%
  Khai Nguyen\\
  Department of Statistics and Data Sciences\\
  The University of Texas at Austin\\
  Austin, TX 78712 \\
  \texttt{khainb@utexas.edu} \\
   \And
   Nhat Ho \\
   Department of Statistics and Data Sciences \\
   The University of Texas at Austin \\
   Austin, TX 78712 \\
   \texttt{minhnhat@utexas.edu} \\
}

\begin{document}

\maketitle

\begin{abstract}
 Seeking informative projecting directions has been an important task in utilizing sliced Wasserstein distance in applications. However, finding these directions usually requires an iterative optimization procedure over the space of projecting directions, which is computationally expensive. Moreover, the computational issue is even more severe in deep learning applications, where computing the distance between two mini-batch probability measures is repeated several times. This nested loop has been one of the main challenges that prevent the usage of sliced Wasserstein distances based on good projections in practice. To address this challenge, we propose to utilize the \textit{learning-to-optimize} technique or \textit{amortized optimization} to predict the informative direction of any given two mini-batch probability measures. To the best of our knowledge, this is the first work that bridges amortized optimization and sliced Wasserstein generative models. In particular, we derive linear amortized models, generalized linear amortized models, and non-linear amortized models which are corresponding to three types of novel mini-batch losses, named \emph{amortized sliced Wasserstein}. We demonstrate the favorable performance of the proposed sliced losses in deep generative modeling on standard benchmark datasets \footnote{Code for the paper  is published at \url{https://github.com/UT-Austin-Data-Science-Group/AmortizedSW}.}.
\end{abstract}

\vspace{-0.5 em}
\section{Introduction}
\vspace{-0.5 em}
\label{sec:introduction}
Generative modeling is one of the most important tasks in machine learning and data science.  Leveraging the expressiveness of neural networks in parameterizing the model distribution, deep generative models such as GANs~\cite{goodfellow2014generative}, VAEs~\cite{kingma2013auto}, and diffusion models~\cite{ho2020denoising,song2019generative}, achieve a significant quality of sampling images. Despite differences in the way of modeling the model distribution, optimization objectives of training generative models can be written as minimizing a discrepancy $\mathcal{D}(\cdot,\cdot)$ between data distribution $\mu$ and the model distribution $\nu_\phi$ with $\phi \in \Phi$, parameter space of neural networks weights, namely,  we solve for $\hat{\phi} \in \argmin_{\phi \in \Phi} \mathcal{D}(\mu,\nu_\phi)$. For example, Kullback–Leibler divergence is used in VAEs and diffusion models, Jensen–Shannon divergence appears in GANs, and f-divergences are utilized in f-GANs~\cite{nowozin2016f}. Because of the complexity of the neural networks $\phi$, closed-form optimal solutions to these optimization problems are intractable. Therefore, gradient-based methods and their stochastic versions are widely used in practice to approximate these solutions. 

Recently, optimal transport-based losses, which we denote as $\mathcal{D}(\cdot,\cdot)$, are utilized to train generative models due to their training stability, efficiency, and geometrically meaning. Examples of these models include Wasserstein GAN~\cite{arjovsky2017wasserstein} with the dual form of Wasserstein-1 distance~\cite{peyre2020computational}, and OT-GANs~\cite{genevay2018learning,salimans2018improving} with the primal form of Wasserstein distance and with Sinkhorn divergence~\cite{cuturi2013sinkhorn} 
 between mini-batch probability measures. Although these models considerably improve the generative performance, there have been remained certain problems. In particular, Wasserstein GAN is reported to fail to approximate the  Wasserstein distance~\cite{stanczuk2021wasserstein} while OT-GAN suffers from high computational complexity of Wasserstein distance: $\mathcal{O}(m^3 \log m)$ and its curse of dimensionality: the sample complexity of $\mathcal{O}(m^{-1/d})$ where $m$ is the number of supports of two mini-batch measures. The entropic regularization~\cite{cuturi2013sinkhorn} had been proposed to improve the computational complexity of approximating optimal transport to $\mathcal{O}(m^2)$~\cite{altschuler2017near, lin2019efficient, Lin-2019-Efficiency, Lin-2020-Revisiting} and to remove the curse of dimensionality~\cite{Mena_2019}. However practitioners usually choose to use the slicing (projecting version) of Wasserstein distance~\cite{wu2019sliced,deshpande2018generative,kolouri2018sliced,nguyen2021improving} due to a fast computational complexity $\mathcal{O}(m \log m)$ and no curse of dimensionality $\mathcal{O}(m^{-1/2})$. The distance is known as sliced Wasserstein distance (SW)~\cite{bonneel2015sliced}. Sliced Wasserstein distance is defined as the expected one-dimensional Wasserstein distance between two projected measures over the uniform distribution over the unit sphere. Due to the intractability of the expectation, Monte Carlo samples from the uniform distribution over the unit sphere are used to approximate the distance. The number of samples is often called the number of projections and it is denoted as $L$.

From applications, practitioners observe that sliced Wasserstein distance requires a sufficiently large number of projections $L$ relative to the dimension of data to perform well~\cite{kolouri2018sliced,deshpande2018generative}. Increasing $L$ leads to a linear increase in computational time and memory. However, when data lie in a low dimensional manifold, several projections are redundant since they collapse projected measures to a Dirac-Delta measure at zero. There are some attempts to overcome that issue including sampling orthogonal directions~\cite{rowland2019orthogonal} and mapping the data to a lower-dimensional space~\cite{deshpande2018generative}. The most popular approach is to search for the direction that maximizes the projected distance, which is known as max-sliced Wasserstein distance ($\maxSW$)~\cite{deshpande2019max}. Nevertheless, in the context of deep generative models and deep learning in general, the optimization over the unit sphere requires iterative projected gradient descent methods that can be computationally expensive. In detail, each gradient-update of the model parameters (neural networks) requires an additional loop for optimization of $\maxSW$ between two mini-batch probability measures. Therefore, we have two nested optimization loops: the global loop (optimizing model parameters) and the local loop (optimizing projection). These optimization loops can slow down the training considerably.

\textbf{Contribution.} 
To overcome the issue, we propose to leverage \textit{learning to learn} techniques (\textit{amortized optimization}) to predict the optimal solution of the local projection optimization. We bridge the literature on amortized optimization and optimal transport by designing amortized models to solve the iterative optimization procedure of finding optimal slices in the sliced Wasserstein generative model.  To the best of our knowledge, this is the first time amortized optimization is used in sliced Wasserstein literature. In summary, our main contributions are two-fold:
\begin{enumerate}
    \item  First, we introduce a novel family of mini-batch sliced Wasserstein losses that utilize amortized models to yield informative projecting directions, named amortized sliced Wasserstein losses ($\ASW$). We specify three types of amortized models: linear amortized, generalized linear amortized, and non-linear amortized models that are corresponding to three mini-batch losses: linear amortized sliced Wasserstein ($\LASW$), generalized linear amortized sliced Wasserstein ($\GASW$), and non-linear amortized sliced Wasserstein ($\NASW$). Moreover, we discuss some properties of $\ASW$ losses including metricity, complexities, and connection to mini-batch Max-SW.

 \item We then introduce the application of $\ASW$ in generative modeling. Furthermore, we carry out extensive experiments on standard benchmark datasets including CIFAR10, CelebA, STL10, and CelebAHQ to demonstrate the favorable performance of $\ASW$ in learning generative models. Finally, we measure the computational speed and memory of $\ASW$, mini-batch Max-SW, and mini-batch SW to show the efficiency of $\ASW$.
\end{enumerate}

\textbf{Organization.} The remainder of the paper is organized as follows. We first provide background about Wasserstein distance, sliced Wasserstein distance, max-sliced Wasserstein distance, and amortized optimization in Section~\ref{sec:background}. In Section~\ref{sec:ASW}, we propose amortized sliced Wasserstein distances and analyze some of their theoretical properties. The discussion on related works is given in Section~\ref{sec:relatedworks}. Section~\ref{sec:experiments} contains the application of $\ASW$ to generative models, qualitative experimental results, and quantitative experimental results on standard benchmarks. In Section~\ref{sec:conclusion}, we provide a conclusion. Finally, we defer the proofs of key results and extra materials to the Appendices.

 \textbf{Notation.} For any $d \geq 2$, $\sphere{d}:=\{\theta \in \mathbb{R}^{d}\mid  ||\theta||_2^2 =1\}$ denotes the $d$ dimensional unit hyper-sphere in $\mathcal{L}_2$ norm, and $\mathcal{U}(\sphere{d})$ is the uniform measure over $\sphere{d}$. Moreover, $\delta$ denotes the Dirac delta function. For $p\geq 1$, $\mathcal{P}_p(\mathbb{R}^d)$ is the set of all probability measures on $\mathbb{R}^d$ that has finite $p$-moments. For $\mu,\nu \in \mathcal{P}_p(\mathbb{R}^d)$, $\Pi(\mu,\nu):=\{\pi \in \mathcal{P}_p(\mathbb{R}^d \times \mathbb{R}^d) \mid \int_{\mathbb{R}^d} \pi(x,y) dx = \nu, \int_{\mathbb{R}^d} \pi(x,y) dy = \mu \}$ is the set of transportation plans between $\mu$ and $\nu$. For $m\geq 1$, we denotes  $\mu^{\otimes m }$ as the product measure which has the supports are the joint vector of $m$ random variables that follows $\mu$. For a vector $X \in \mathbb{R}^{dm}$, $X:=(x_1,\ldots,x_m)$, $P_{X}$ denotes the empirical measures $\frac{1}{m} \sum_{i=1}^m \delta_{x_i}$. We denote $\theta \sharp \mu$ as the push-forward  probability measure of $\mu$ through the function $T_\theta: \mathbb{R}^d \to \mathbb{R}$ where  $T_\theta(x) = \theta^\top x$.

\section{Background}

\label{sec:background}
In this section, we first review the definitions of the Wasserstein distance, the sliced Wasserstein distance, and the max-sliced Wasserstein distance. We then formulate generative models based on the max-sliced Wasserstein distances and review the amortized optimization problem and its application to the max-sliced Wasserstein generative models.

\subsection{(Sliced)-Wasserstein Distances}

\label{sec:Wasserstein_def}
We first define the Wasserstein-$p$ distance~\cite{Villani-09, peyre2019computational} between two probability measures $\mu \in \mathcal{P}_p(\mathbb{R}^d)$ and $\nu \in \mathcal{P}_p(\mathbb{R}^d)$ as follows: $\text{W}_p(\mu,\nu) : = \Big{(} \inf_{\pi \in \Pi(\mu,\nu)} \int_{\mathbb{R}^d \times \mathbb{R}^d} \| x - y\|_p^{p} d \pi(x,y) \Big{)}^{\frac{1}{p}}$. When $d=1$, the Wasserstein distance has a closed form which is $W_p(\mu,\nu) =
    ( \int_0^1 |F_\mu^{-1}(z) - F_{\nu}^{-1}(z)|^{p} dz )^{1/p}$
where $F_{\mu}$ and $F_{\nu}$  are  the cumulative
distribution function (CDF) of $\mu$ and $\nu$ respectively. 

To utilize this closed-form property of Wasserstein distance in one dimension and overcome the curse of dimensionality of Wasserstein distance in high dimension, the sliced Wasserstein distance~\cite{bonneel2015sliced} between $\mu$ and $\nu$ had been introduced and admitted the following formulation: $\text{SW}_p(\mu,\nu) : = \left(\int_{\mathbb{S}^{d-1}} \text{W}_p^p (\theta \sharp \mu,\theta \sharp \nu) d\theta \right)^{\frac{1}{p}}$.
For each $\theta \in \mathbb{S}^{d- 1}$, $\text{W}_p^p (\theta \sharp \mu,\theta \sharp \nu)$ can be computed in linear time $\mathcal{O}(n \log n)$ where $n$ is the number of supports of $\mu$ and $\nu$. However, due to the integration over the unit sphere, the sliced Wasserstein distance does not have closed-form expression. To approximate the intractable expectation, Monte Carlo scheme is used, namely, we draw uniform samples $\theta_1,\ldots,\theta_L \sim \mathcal{U}(\sphere{d})$ from the unit sphere and obtain the following approximation: $\text{SW}_p (\mu,\nu) \approx  \left(\frac{1}{L}\sum_{i=1}^L \text{W}_p^p (\theta_i \sharp \mu,\theta_i \sharp \nu) \right)^{\frac{1}{p}}$.
In practice, $L$ should be chosen to be sufficiently large compared to the dimension $d$. It is not appealing since the computational complexity of SW is linear with $L$. To reduce projection complexity, max-sliced Wasserstein (Max-SW) is introduced~\cite{deshpande2019max} . In particular, the max-sliced Wasserstein distance between $\mu$ and $\nu$ is given by:
\begin{align} 
    \maxSW(\mu,\nu) := \max_{\theta \in \sphere{d}} \text{W}_p (\theta \sharp \mu,\theta \sharp \nu).
\end{align}
To solve the optimization problem, a projected gradient descent procedure is used. We present a simple algorithm in Algorithm~\ref{alg:MaxSW}.  In practice, practitioners often set a fixed number of gradient updates, e.g., $T=100$.

\subsection{Learning Generative Models with Max-Sliced Wasserstein and Amortized Optimization}
We now provide an application of (sliced)-Wasserstein distances to generative models settings. The problem can be seen as the following optimization:
\begin{align}
    \min_{\phi \in \Phi} \mathcal{D}(\mu,\nu_\phi),
\end{align}
where $\mathcal{D}(\cdot,\cdot)$ can be Wasserstein distance or SW distance or Max-SW distance.
Despite the recent progress on scaling up Wasserstein distance in terms of the size of supports of probability measures~\cite{altschuler2017near, lin2019efficient}, using the original form of Wasserstein distances is still not tractable in real training due to both the memory constraint and time constraint. In more detail, the number of training samples is often huge, e.g., one million, and the dimension of data is also huge ,e.g., ten thousand.  Therefore, mini-batch losses based on Wasserstein distances have been proposed~\cite{fatras2020learning, nguyen2022transportation, nguyen2022improving}. The corresponding population form of these losses between two probability measures $\mu$ and $\nu$ is:
\begin{align}
    \tilde{\mathcal{D}}(\mu,\nu):= \mathbb{E}_{X,Y \sim \mu^{\otimes m}\otimes \nu^{\otimes m}} \mathcal{D}(P_X,P_Y), \label{eq:pop_minibatch_Wasserstein}
\end{align}
where $m\geq 1$ is the mini-batch size and $\mathcal{D}$ is a Wasserstein metric. 

In the generative model context~\cite{goodfellow2014generative}, a stochastic gradient of the parameters of interest is utilized to update these parameters, namely, 
\begin{align}
    \nabla_\phi  \tilde{\mathcal{D}}(\mu,\nu_\phi) \approx \frac{1}{k} \sum_{i=1}^k \nabla_\phi \mathcal{D}(P_{X_i},P_{Y_{\phi_i}}),
\end{align}
where $k$ is the number of mini-batches (is often set to 1), and $(X_i,Y_{\phi_i})$ is i.i.d sample from $\mu^{\otimes m}\otimes \nu_\phi^{\otimes m}$. The exchangeability between derivatives and expectation, and unbiasedness of the stochastic gradient are proven in~\cite{fatras2021minibatch}. Mini-batch losses are not distances; however, we can derive mini-batch energy distances from them~\cite{salimans2018improving}.

\begin{algorithm}[!t]
\caption{Max-sliced Wasserstein distance}
\begin{algorithmic}
\label{alg:MaxSW}
\STATE \textbf{Input:} Probability measures: $\mu,\nu$, learning rate $\eta$, max number of iterations $T$.
  \STATE Initialize $\theta$
  \WHILE{$\theta$ not converge or reach $T$}
  \STATE $\theta = \theta +  \eta \cdot \nabla_\theta\text{W}_p (\theta \sharp \mu,\theta \sharp \nu)$
  \STATE $\theta = \frac{\theta}{||\theta||_2}$
  \ENDWHILE
 \STATE \textbf{Return:} $\theta$
\end{algorithmic}
\end{algorithm}
\textbf{Learning generative models via max-sliced Wasserstein:} As we mentioned in Section~\ref{sec:Wasserstein_def}, the max-sliced Wasserstein distance can overcome the curse of dimensionality of the Wasserstein distance and the issues of Monte Carlo samplings in the sliced Wasserstein distance. Therefore, it is an appealing divergence for learning generative models. By replacing the Wasserstein metric in equation~(\ref{eq:pop_minibatch_Wasserstein}), we arrive at the following formulation of the mini-batch max-sliced Wasserstein loss, which is given by:
\begin{align}
\label{eq:mmaxsw}
    \text{m-}\maxSW(\mu,\nu) = \mathbb{E}_{X,Y \sim \mu^{\otimes m}\otimes \nu^{\otimes m}} \left[\max_{\theta \in \sphere{d}} \text{W}_p (\theta \sharp P_X,\theta \sharp P_Y)\right].
\end{align}
Here, we can observe that each pair of mini-batch contains its own optimization problem of finding the "max" slice. Placing this in the context of iterative training of generative models, we can foresee its expensive computation. For a better understanding, we present an algorithm for training generative models with mini-batch max-sliced Wasserstein in Algorithm~\ref{alg:trainingMaxSW}. In practice, there are some modifications of training generative models with mini-batch Max-SW for dealing with unknown metric space~\cite{deshpande2018generative}. We defer the details of these modifications in Appendix~\ref{sec:training_detail}.

\begin{algorithm}[!t]
\caption{Training generative models with mini-batch max-sliced Wasserstein loss}
\begin{algorithmic}
\label{alg:trainingMaxSW}
\STATE \textbf{Input:} Data probability measure $\mu$, model learning rate $\eta_1$, slice learning rate $\eta_2$, model maximum number of iterations $T_1$,  slice maximum number of iterations $T_2$, number of mini-batches $k$ (is often set to 1).
  \STATE Initialize $\phi$, the model probability measure $\nu_\phi$
  \WHILE{$\phi$ not converge or reach $T_1$}
  \STATE $\nabla_\phi = 0$
  \STATE Sample $(X_1,Y_{\phi,1}),\ldots,(X_k,Y_{\phi,k}) \sim \mu^{\otimes m}\otimes \nu_\phi^{\otimes m}$
  \FOR{$i=1$ to $k$}
   \WHILE{$\theta$ not converge or reach $T_2$}
  \STATE $\theta = \theta +  \eta_2 \cdot \nabla_\theta\text{W}_p (\theta \sharp P_{X_i},\theta \sharp P_{Y_{\phi,i}})$
  \STATE $\theta = \frac{\theta}{||\theta||_2}$
  \ENDWHILE
  \STATE $\nabla_\phi = \nabla_\phi + \frac{1}{k} \nabla_\phi \text{W}_p (\theta \sharp P_{X_i},\theta \sharp P_{Y_{\phi,i}})$
  \ENDFOR
  \STATE $\phi = \phi - \eta_1 \cdot \nabla_\phi$
  \ENDWHILE
 \STATE \textbf{Return:} $\phi,\nu_\phi$
\end{algorithmic}
\end{algorithm}

\textbf{Amortized optimization:} A natural question appears: "How can we avoid the nested loop in mini-batch Max-SW due to several local optimization problems?". In this paper, we propose a practical solution for this problem, which is known as \textit{amortized optimization}~\cite{amos2022tutorial}. In amortized optimization, instead of solving all optimization problems independently, an amortized model is trained to predict optimal solutions to all problems. We now state the adapted definition of amortized models based on that in~\cite{ruishu2017,amos2022tutorial}:

\begin{definition}
\label{def:amodel} For each context variable $x$ in the context space $\mathcal{X}$, $\theta^\star(x)$ is the solution of the optimization problem $\theta^\star(x) = \argmin_{\theta \in \Theta} \mathcal{L}(\theta,x)$, where $\Theta$ is the solution space. A parametric function $f_\psi: \mathcal{X} \to \Theta$, where $\psi \in \Psi$, is called an amortized model if 
\begin{align}
\label{eq:famortized}
    f_\psi (x) \approx \theta^\star (x), \quad \forall x \in \mathcal{X}.
\end{align}
The amortized model is trained by the amortized optimization objective which is defined as:
\begin{align}
\label{eq:amortizedobjective}
\min_{\psi \in \Psi} \mathbb{E}_{x \sim p(x)} \mathcal{L}(f_\psi(x),x),
\end{align}
where  $p(x)$ is a probability measure on $\mathcal{X}$ which measures the "importance" of optimization problems.
\end{definition}

The amortized model in Definition~\ref{def:amodel} is sometimes called a \textit{fully} amortized model for a distinction with the other concept of \textit{semi} amortized model~\cite{amos2022tutorial}. The gap between the predicted solution and the optimal solution $\mathbb{E}_{x\sim p(x)}||f_\psi (x) - \theta^\star(x)||_2 $ is called  the amortization gap. However, understanding this gap depends on specific configurations of the objective $\mathcal{L}(\cdot, x)$, such as convexity and smoothness, which are often non-trivial to obtain in practice.

\section{Amortized Sliced Wasserstein}

\label{sec:ASW}
In this section, we discuss an application of amortized optimization to the mini-batch max-sliced Wasserstein. In particular, we first formulate the approach into a novel family of mini-batch losses, named \emph{Amortized Sliced Wasserstein}. Each member of this family utilizes an amortized model for predicting informative slicing directions of mini-batch measures. We then propose several useful amortized models in practice, including the linear model, the generalized linear model, and the non-linear model.

\subsection{Amortized Sliced Wasserstein and Amortized Models}

\label{subsec:ASW}
We extend the definition of the mini-batch max-sliced Wasserstein in Equation~(\ref{eq:mmaxsw}) with the usage of an amortized model to obtain the amortized sliced Wasserstein as follows.

\begin{definition}
\label{def:asw}
Let $p\geq 1$, $m \geq 1$, and $\mu,\nu$ are two probability measures in $\mathcal{P}(\mathbb{R}^d)$. Given an amortized model $f_\psi:\mathbb{R}^{dm}  \times \mathbb{R}^{dm} \to \sphere{d}$ where $\psi \in \Psi$, the \emph{amortized sliced Wasserstein} between $\mu$ and $\nu$ is:
\begin{align}
    \ASW(\mu,\nu):= \max_{\psi \in \Psi}\mathbb{E}_{(X,Y) \sim \mu^{\otimes m} \otimes \nu^{\otimes m}}[\text{W}_p(f_\psi (X,Y) \sharp P_X,f_\psi (X,Y) \sharp P_Y)].
\end{align}
\end{definition}
From the definition, we can see that the amortized model maps each pair of mini-batches to the optimal projecting direction on the unit hypersphere between two corresponding mini-batch probability measures. We have the following result about the symmetry and positivity of $\ASW$.
\begin{proposition}
\label{proposition:amortized_sliced}
The  amortized sliced Wasserstein losses are positive and symmetric. However, they are not metrics since they do not satisfy the identity property, namely, $\ASW(\mu,\nu) = 0 \notiff \mu = \nu$.
\end{proposition}
Proof of Proposition~\ref{proposition:amortized_sliced} is in Appendix~\ref{subsec:proof:proposition:amortized_sliced}. Our next result indicates that we can upper bound the amortized sliced Wasserstein in terms of mini-batch max-sliced Wasserstein.
\begin{proposition}
\label{prop:lower_bound}
Assume that the space $\Psi$ is a compact set and the function $f_{\psi}$ is continuous in terms of $\psi$. Then, the amortized sliced Wasserstein are lower-bounds of the mini-batch max-sliced Wasserstein (Equation~\ref{eq:mmaxsw}), i.e.,  $\ASW(\mu,\nu)  \leq  \text{m-Max-SW}(\mu, \nu)$ for all probability measures $\mu$ and $\nu$.
\end{proposition} 
Proof of Proposition~\ref{prop:lower_bound} is in Appendix~\ref{subsec:proof:lower_bound}.

\textbf{Parametric forms of the amortized model:} Now we define three types of amortized models that we will use in the experiments.
\begin{definition}
\label{def:linearmodel} Given $X,Y \in  \mathbb{R}^{dm}$, and the one-one "reshape" mapping $T:\mathbb{R}^{dm} \to \mathbb{R}^{d \times m}$, the \emph{linear amortized model} is defined as:
\begin{align}
    f_\psi (X,Y) := \frac{w_0+T(X)w_1 + T(Y)w_2}{||w_0+T(X)w_1 + T(Y)w_2 ||_2^2},
\end{align}
where $w_1,w_2 \in \mathbb{R}^{ m}$, $w_0 \in \mathbb{R}^d $ and $\psi =(w_0,w_1,w_2)$.
\end{definition}

In Definition~\ref{def:linearmodel}, the assumption is that the optimal projecting direction lies on the subspace that is spanned by the basis $\{x_1,\ldots,x_m,y_1,\ldots,y_m,w_0\}$ where $X=(x_1,\ldots,x_m)$ and $Y=(y_1,\ldots,y_m)$. The computational complexity of this function is $\mathcal{O}((2m+1)d)$ since those of the operators $T(X)w_1$ and $T(Y)w_2$ are $\mathcal{O}(md)$ while adding the bias $w_0$ costs an additional computational complexity $\mathcal{O}(d)$. The number of parameters in linear amortized model is $2m+d$. 

To increase the expressiveness of the linear amortized model, we apply some (non-linear) mappings to the inputs $X$ and $Y$, which results in the generalized linear amortized model as follows.

\begin{definition}
\label{def:glinearmodel} Given $X,Y \in  \mathbb{R}^{dm}$, and the one-one "reshape" mapping $T:\mathbb{R}^{dm} \to \mathbb{R}^{d \times m}$, the \emph{generalized linear amortized model} is defined as:
\begin{align}
    f_\psi (X,Y) := \frac{w_0+ T(g_{\psi_1}(X)) w_1 +  T(g_{\psi_1}(Y))w_2}{||w_0+ T(g_{\psi_1}(X)) w_1 + T(g_{\psi_1}(Y)) w_2||_2^2},
\end{align}
where $w_1,w_2 \in \mathbb{R}^{m}$, $w_0 \in \mathbb{R}^d $,  $\psi_1 \in \Psi_1$, $g_{\psi_1}: \mathbb{R}^{dm} \to \mathbb{R}^{dm}$ and $\psi =(w_0,w_1,w_2,\psi_1)$.
\end{definition}

In Definition~\ref{def:glinearmodel}, the assumption is that the optimal projecting direction lies on the subspace that is spanned by the basis $\{x'_1,\ldots,x'_m,y'_1,\ldots,y'_m,w_0\}$ where $g_{\psi_1}(X)=(x'_1,\ldots,x'_m)$ and $g_{\psi_1}(Y)=(y'_1,\ldots,y'_m)$. To specify, we let $g_{\psi_1}(X) = (W_2\sigma(W_1 x_1)+b_0 ,\ldots, W_2\sigma(W_1 x_m) +b_0) $, where $\sigma(\cdot)$ is the Sigmoid function, $W_1 \in \mathbb{R}^{d\times d}$, $W_2 \in \mathbb{R}^{d\times d}$, and $b_0 \in \mathbb{R}^d$. Compared to the linear model, the generalized linear model needs additional computations for $g_\psi (T(X))$ and $ g_\psi(T(Y))$, which are at the order of $\mathcal{O}(2m(d^2+d))$. It is because we need to include the complexity for matrix multiplication, e.g., $W_1 x_1$ that costs $\mathcal{O}(d^2)$, for Sigmoid function that costs $\mathcal{O}(d)$, and for adding bias $b_0$ that costs $\mathcal{O}(d)$. Therefore, the total computational complexity of the function $f_\psi$ is $\mathcal{O}(4md^2 + 6md + d)$ while the number of parameters is $2(m+d^2+d)$.

We finally propose another amortized model where we instead consider some mapping on the function $\omega_{0} + T(X) \omega_{1} + T(Y) \omega_{2}$ in the linear amortized model so as to increase the approximation power of the function $f_{\psi}$.

\begin{definition}
\label{def:nonlinearmodel} Given $X,Y \in  \mathbb{R}^{dm}$, and the one-one "reshape" mapping $T:\mathbb{R}^{dm} \to \mathbb{R}^{d \times m}$, the \emph{non-linear amortized model} is defined as:
\begin{align}
    f_\psi (X,Y) := \frac{ h_{\psi_2}(w_0+ T(X)w_1 + T(Y)w_2)}{||h_{\psi_2}(w_0+T(X)w_1 + T(Y)w_2)||_2^2},
\end{align}
where $w_1,w_2 \in \mathbb{R}^{m}$, $w_0 \in \mathbb{R}^d $, $\psi_2 \in \Psi_2$, $h_{\psi_2}:\mathbb{R}^d \to \mathbb{R}^d$ and $\psi =(w_0,w_1,w_2,\psi_2)$.
\end{definition}

In Definition~\ref{def:nonlinearmodel}, the assumption is that the optimal projecting direction lies on the image of the function $h_{\psi_2}(\cdot)$ that maps from the subspace spanned by $\{x_1,\ldots,x_m,y_1,\ldots,y_m,w_0\}$ where $X=(x_1,\ldots,x_m)$ and $Y=(y_1,\ldots,y_m)$. The computational complexity for $h_{\psi_2}(x) = W_4 \sigma(W_3x)) +b_0 $ when $x \in \mathbb{R}^d$, $W_3 \in \mathbb{R}^{d\times d}$,  $W_4 \in \mathbb{R}^{d\times d}$, and $b_0 \in \mathbb{R}^d$ is at the order of $\mathcal{O}(2(d^2+d))$. Therefore, the total computational complexity of the function $f_\psi$ is $\mathcal{O}(2md+ 2d^2+3d)$ while the number of parameters is $2(m+d^2+d)$.

Using amortized models in Definitions~\ref{def:linearmodel}-\ref{def:nonlinearmodel} leads to three amortized sliced Wasserstein losses, which are linear amortized sliced Wasserstein loss ($\LASW$),  generalized linear amortized sliced Wasserstein loss ($\GASW$), and non-linear amortized sliced Wasserstein loss ($\NASW$) in turn.

\begin{remark}
The parametric forms in Definitions~\ref{def:linearmodel}-\ref{def:nonlinearmodel} are chosen as they are well-known choices for parametric functions. There are still several other ways of parameterization that can be utilized in practice based on prior knowledge about data, e.g., we can use convolution operator for saving parameters or we can strengthen the dependence between samples via recursive functions. We leave the design of these amortized models for future work.
\end{remark}


\subsection{Amortized Sliced Wasserstein Generative Models}
\label{subsec:ASWgen}
Based on the amortized sliced Wasserstein losses, our objective function for training a generative model $\nu_\phi$ parametrized by $\phi \in \Phi$ now becomes:
\begin{align*}
     \min_{\phi \in \Phi }\max_{\psi \in \Psi}\mathbb{E}_{(X,Y_\phi ) \sim \mu^{\otimes m} \otimes \nu_\phi^{\otimes m}}[\text{W}_p(f_\psi (X,Y_\phi) \sharp P_X,f_\psi (X,Y_\phi) \sharp P_{Y_\phi}) ] :=  \min_{\phi \in \Phi }\max_{\psi \in \Psi} \mathcal{L}(\mu,\nu_\phi, \psi). 
\end{align*}
Since the above optimization forms a minimax problem, we can use an alternating stochastic gradient descent-ascent algorithm to solve it. In particular, the stochastic gradients of $\phi$ and $\psi$ can be estimated from mini-batches $(X_1,Y_{\phi,1}) ,\ldots, (X_k,Y_{\phi,k}) \sim \mu^{\otimes m} \otimes \nu_\phi^{\otimes m}$ as follows:
\vspace{-0.5 em}
\begin{align}
    \nabla_\phi  \mathcal{L}(\mu,\nu_\phi,\psi) =  \frac{1}{k}\sum_{i=1}^k \nabla_\phi \text{W}_p (f_\psi(X_i,Y_{\phi,i}) \sharp P_{X_i},f_\psi(X_i,Y_{\phi,i}) \sharp P_{Y_{\phi,i}}), \\
    \nabla_\psi \mathcal{L}(\mu,\nu_\phi,\psi) =  \frac{1}{k}\sum_{i=1}^k  \nabla_\psi  \text{W}_p (f_\psi(X_i,Y_{\phi,i}) \sharp P_{X_i},f_\psi(X_i,Y_{\phi,i}) \sharp P_{Y_{\phi,i}}). 
\end{align}
For more details, we present the procedure in Algorithm~\ref{alg:trainingASW}. 

\begin{algorithm}[!t]
\caption{Training generative models with amortized sliced Wasserstein loss}
\begin{algorithmic}
\label{alg:trainingASW}
\STATE \textbf{Input:} Data probability measure $\mu$, model learning rate $\eta_1$, amortized learning rate $\eta_2$, maximum number of iterations $T$, number of mini-batches $k$ (is often set to 1).
  \STATE Initialize $\phi$, the model probability measure $\nu_\phi$.
  \STATE Initialize $\psi$, the amortized model $f_\psi$.
  \WHILE{$\phi,\psi$ not converge or reach $T$}
  \STATE $\nabla_\phi = 0; \nabla_\psi = 0$
  \STATE Sample $(X_1,Y_{\phi,1}),\ldots,(X_k,Y_{\phi,k}) \sim \mu^{\otimes m}\otimes \nu_\phi^{\otimes m}$
  \FOR{$i=1$ to $k$}
  \STATE $\nabla_\phi = \nabla_\phi + \frac{1}{k} \nabla_\phi \text{W}_p (f_\psi(X_i,Y_{\phi,i}) \sharp P_{X_i},f_\psi(X_i,Y_{\phi,i}) \sharp P_{Y_{\phi,i}})$
  \STATE $\nabla_\psi = \nabla_\psi + \frac{1}{k} \nabla_\psi \text{W}_p (f_\psi(X_i,Y_{\phi,i}) \sharp P_{X_i},f_\psi(X_i,Y_{\phi,i}) \sharp P_{Y_{\phi,i}})$
  \ENDFOR
  \STATE $\phi = \phi - \eta_1\cdot \nabla_\phi$
  \STATE $\psi = \psi + \eta_2\cdot \nabla_\psi$
  \ENDWHILE
 \STATE \textbf{Return:} $\phi,\nu_\phi$
\end{algorithmic}
\end{algorithm}

\textbf{Computational complexity:} From Algorithm~\ref{alg:trainingMaxSW} and Algorithm~\ref{alg:trainingASW}, we can see that training with $\ASW$ can escape the inner while-loop for finding the optimal projecting directions. In each iteration of the global while-loop, the computational complexity of computing the mini-batch $\maxSW$ is $\mathcal{O}(2kT_2 (m \log m + dm))$, which is composed by $k$ mini-batches with $T_2$ loops of the projection to one-dimension operator which costs $\mathcal{O}(2dm)$ and the computation of the sliced Wasserstein which costs $\mathcal{O}(2m\log m)$. For the  mini-batch sliced Wasserstein, the overall computational complexity is $\mathcal{O}(2kL(m\log m + dm))$ where $L$ is the number of projections. For $\LASW$,  the overall computation complexity is $\mathcal{O}(2k(m\log m + 3md+d))$ where the extra complexity $\mathcal{O}((2m+1)d)$ comes from the computation of $f_\psi(\cdot)$ (see Section~\ref{subsec:ASW}). Similarly, the computational complexities of $\GASW$ and $\NASW$ are respectively $\mathcal{O}(2k(m\log m + 4md^2 +7md+d))$ and $\mathcal{O}(2k(m\log m + 3md+2d^2+3d))$. 

\textbf{Projection Complexity:} Compared to the sliced Wasserstein, $\maxSW$ reduces the space for projecting directions from $\mathcal{O}(L)$ to $\mathcal{O}(1)$. For $\LASW$, $\GASW$, and $\NASW$, the projection complexity is also $\mathcal{O}(1)$. However, compared to $d$ parameters of Max-SW, $\LASW$  needs $2m+d$ parameters for creating the projecting directions while $\GASW$ and $\NASW$ respectively need $\mathcal{O}(2(m+d^2+d))$ parameters for producing the directions (see Section~\ref{subsec:ASW}). 

\begin{remark}

The computational complexities and the projection complexities of $\GASW$ and $\NASW$ are based on the specific parameterization that we choose in Section~\ref{sec:ASW}. We would like to recall that these complexities can be reduced by lighter parameterization as in the remark at the end of Section~\ref{subsec:ASW}. 
\end{remark}

\section{Related Works}

\label{sec:relatedworks}
Generalized sliced Wasserstein~\cite{kolouri2019generalized} was introduced by changing the push-forward function from linear $T_\theta (x) = \theta^\top x$ to non-linear $T_\theta (x) =g( \theta, x)$ for some non-linear function $g(\cdot,\cdot)$. To cope with the projection complexity of sliced Wasserstein, a biased approximation based on the concentration of Gaussian projections was proposed in~\cite{nadjahi2021fast}. An implementation technique that utilizes both RAM and GPUs' memory for training sliced Wasserstein generative model was introduced in~\cite{lezama2021run}. Augmenting the data to a higher-dimensional space for a better linear separation results in augmented sliced Wasserstein~\cite{chen2022augmented}.  Projected Robust Wasserstein (PRW) metrics appeared in~\cite{paty2019subspace} that finds the best orthogonal linear projecting operator onto $d'>1$ dimensional space. Riemannian optimization techniques for solving PRW were proposed in~\cite{lin2020projection,huang2021riemannian}. We would like to recall that, amortized optimization techniques can be also applied to the case of PRW, max-K-sliced Wasserstein~\cite{dai2021sliced}, sliced divergences~\cite{nadjahi2020statistical}, and might be applicable for sliced mutual information~\cite{goldfeld2021sliced}. Statistical guarantees of training generative models with sliced Wasserstein were derived in~\cite{nadjahi2019asymptotic}.
 
Amortized optimization was first introduced in the form of amortized variational inference~\cite{kingma2013auto,rezende2014stochastic}. Several techniques were proposed to improve the usage of amortized variational inference such as using meta sets in~\cite{wu2020meta}, using iterative amortized variational inference in~\cite{marino2018iterative}, using regularization in~\cite{shu2018amortized}. Amortized inference was also applied into many applications such as probabilistic reasoning~\cite{gershman2014amortized}, probabilistic programming~\cite{ritchie2016deep}, and structural learning~\cite{chang2015structural}. However, to the best of our knowledge, it is the first time that amortized optimization is used in the literature of optimal transport. We refer to~\cite{amos2022tutorial} for a tutorial about the amortized optimization.

\section{Experiments}

\label{sec:experiments}
In this section, we focus on comparing $\ASW$ generative models with SNGAN~\cite{miyato2018spectral}, the sliced Wasserstein generator~\cite{deshpande2018generative}, and the max-sliced Wasserstein generator~\cite{deshpande2019max}. The parameterization of model distribution is based on the neural network architecture of SNGAN~\cite{miyato2018spectral}. The detail of the training processes of all models is given in Appendix~\ref{sec:training_detail}. For datasets, we choose standard benchmarks such as CIFAR10 (32x32)~\cite{krizhevsky2009learning}, STL10 (96x96)~\cite{coates2011analysis}, CelebA (64x64), and CelebAHQ (128x128)~\cite{liu2015faceattributes}.
For quantitative comparison, we use the FID score~\cite{heusel2017gans} and the Inception score (IS)~\cite{salimans2016improved}. We also show some randomly generated images from different models for qualitative comparison. We give full experimental results in Appendix~\ref{sec:fullexp}. The detailed settings about architectures, hyperparameters, and evaluation of FID and IS are given in Appendix~\ref{sec:settings}. We would like to recall that all losses that are used in this section are in their mini-batch version.

 \begin{figure*}[!t]
\begin{center}
    
  \begin{tabular}{cccc}
  \widgraph{0.23\textwidth}{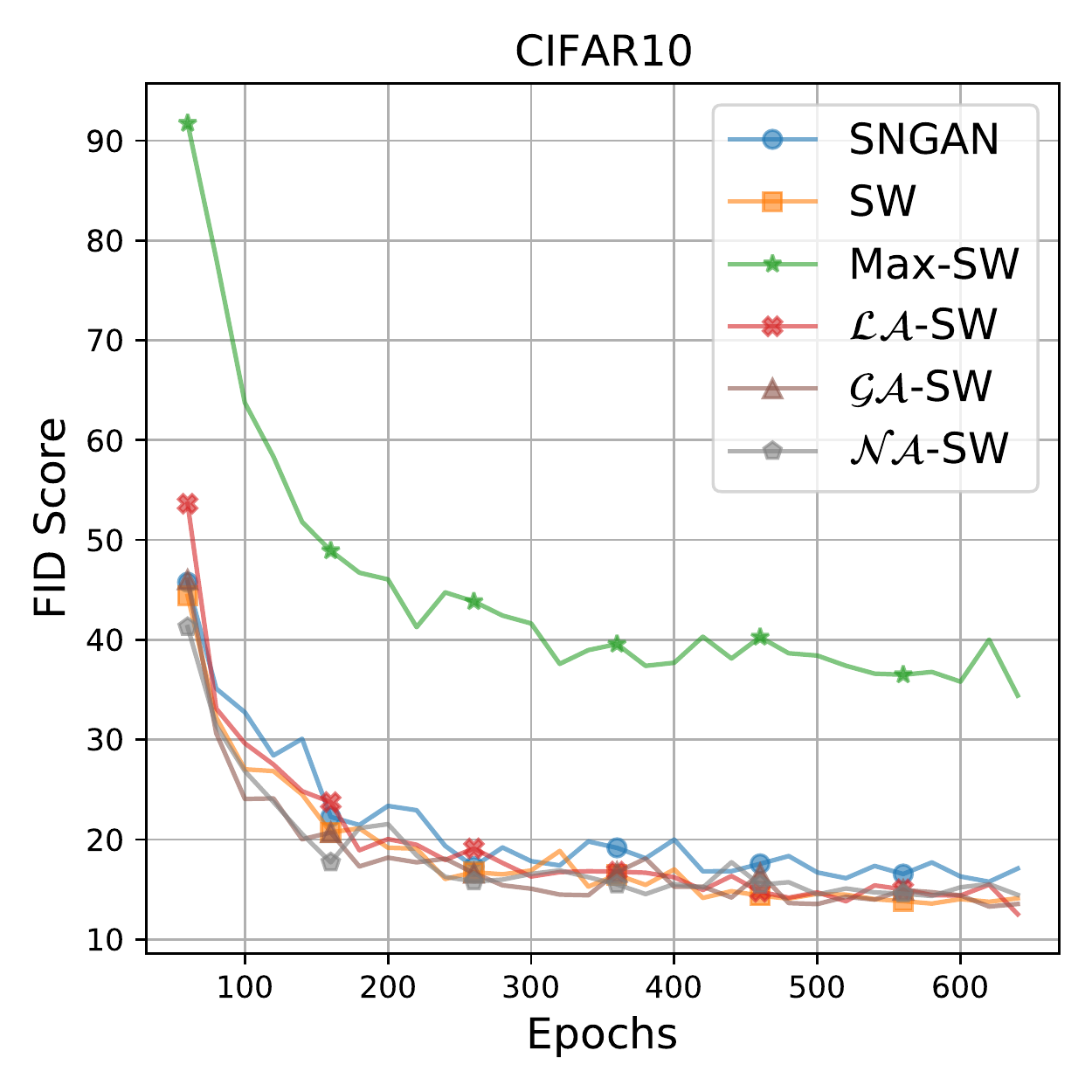} 
  &
\widgraph{0.23\textwidth}{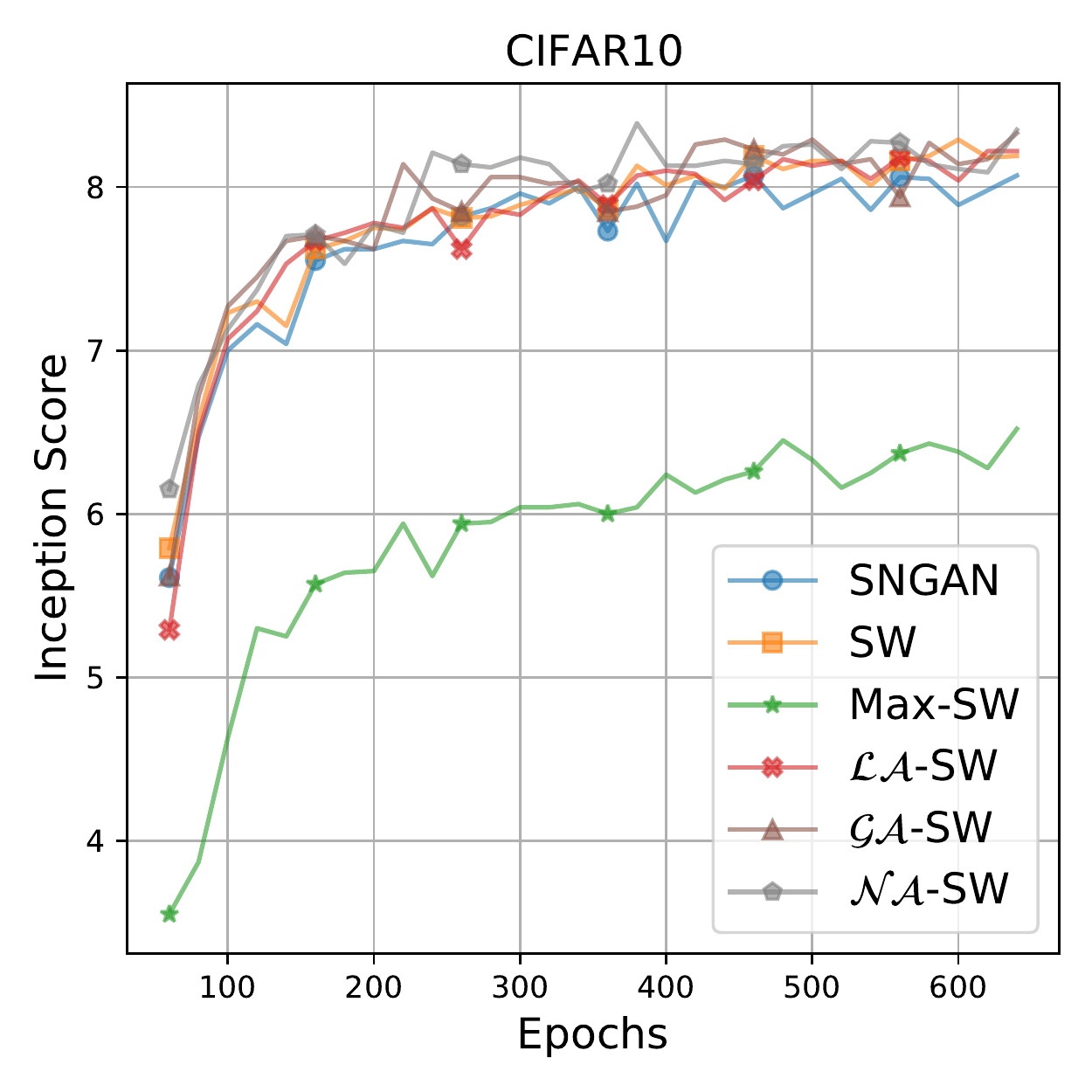} 
&
\widgraph{0.23\textwidth}{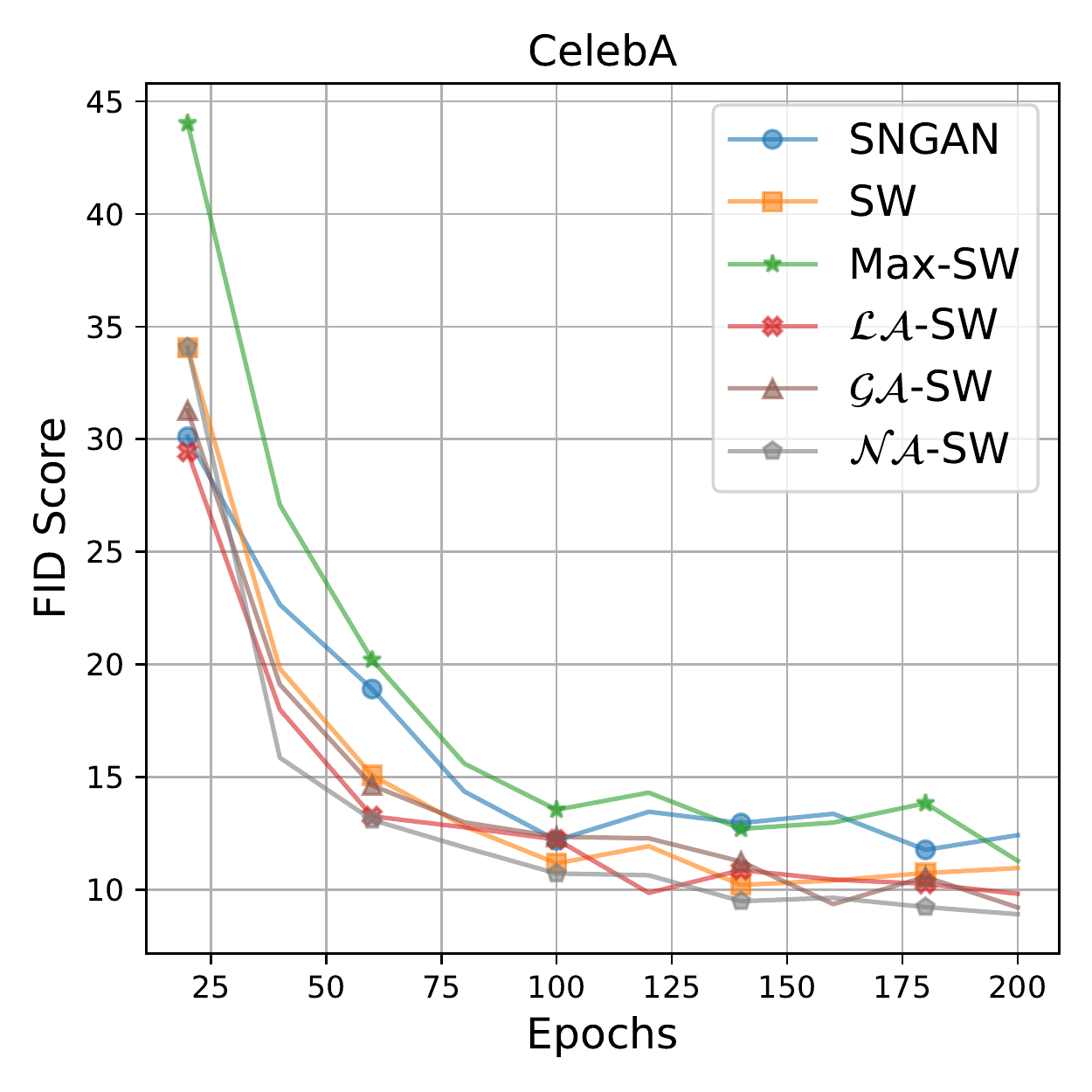} 
  &
\widgraph{0.23\textwidth}{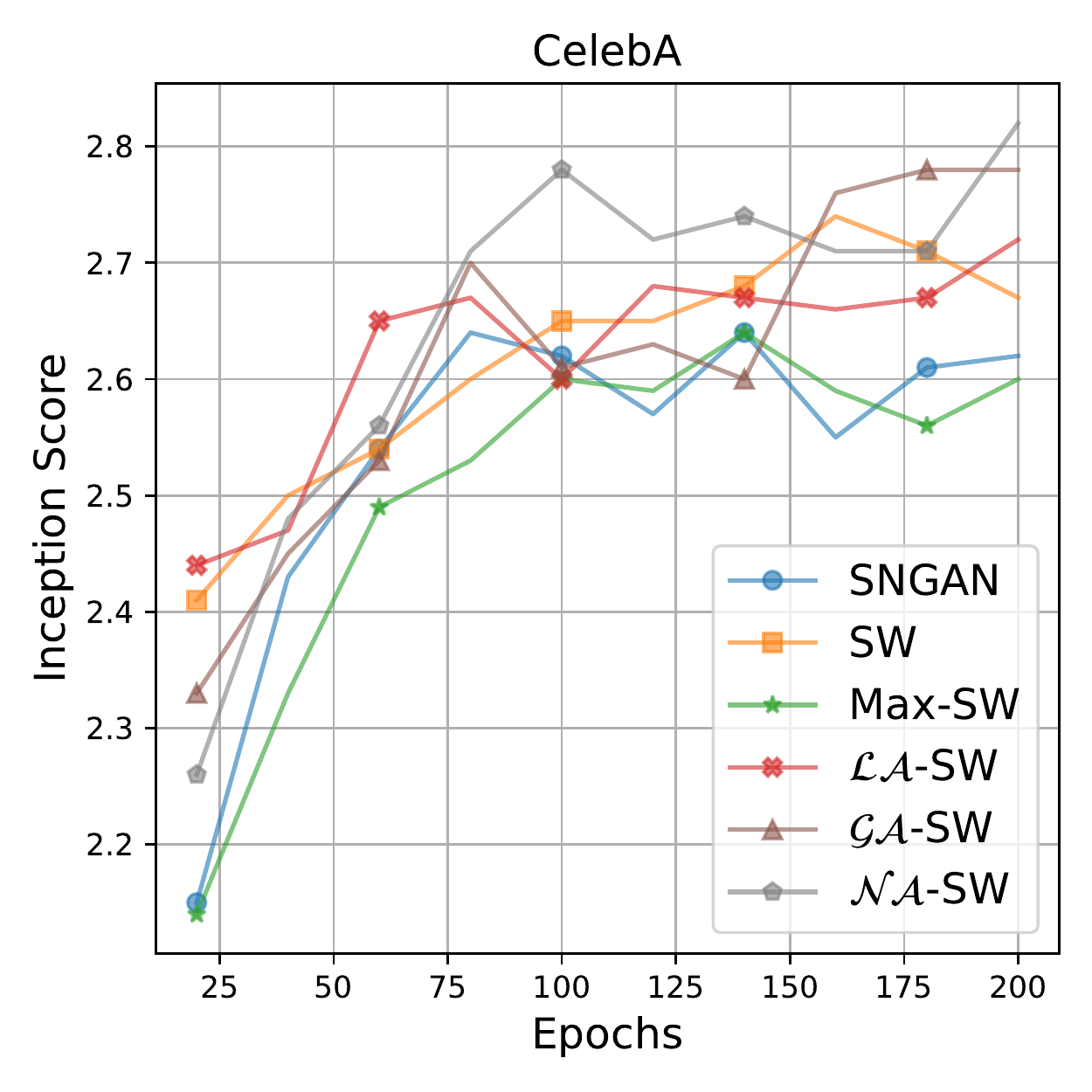} 
\\
\widgraph{0.23\textwidth}{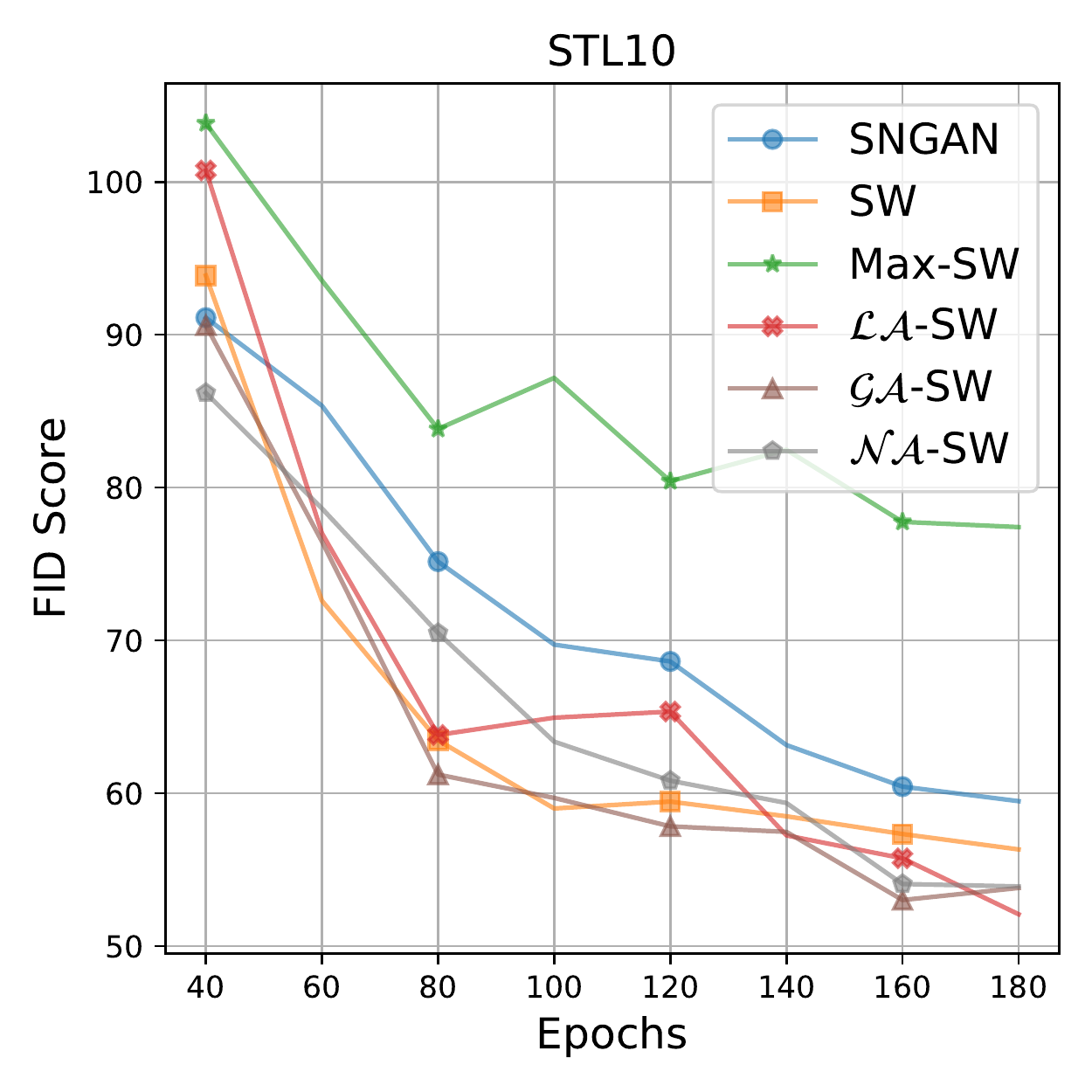} 
  &
\widgraph{0.23\textwidth}{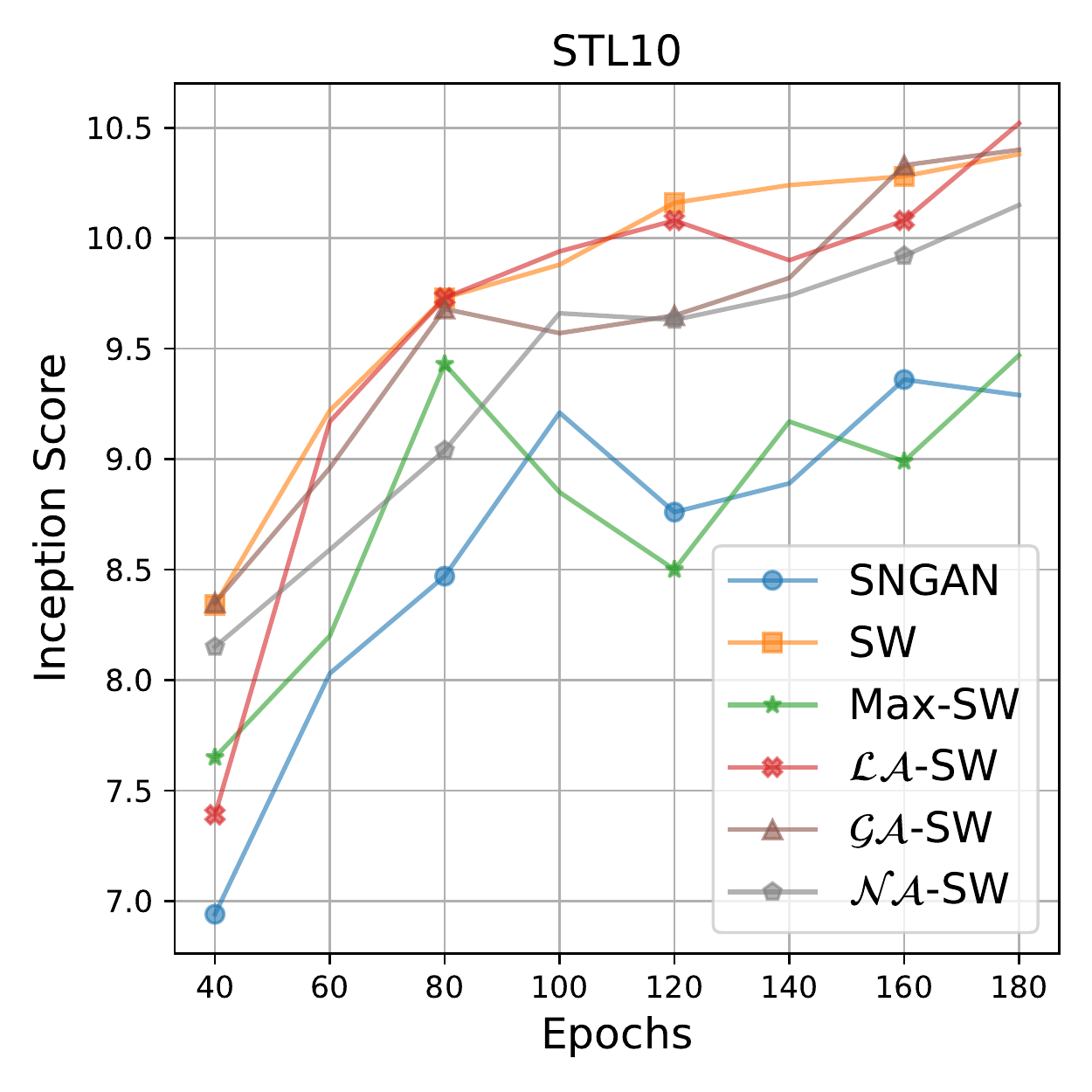} 
&
\widgraph{0.23\textwidth}{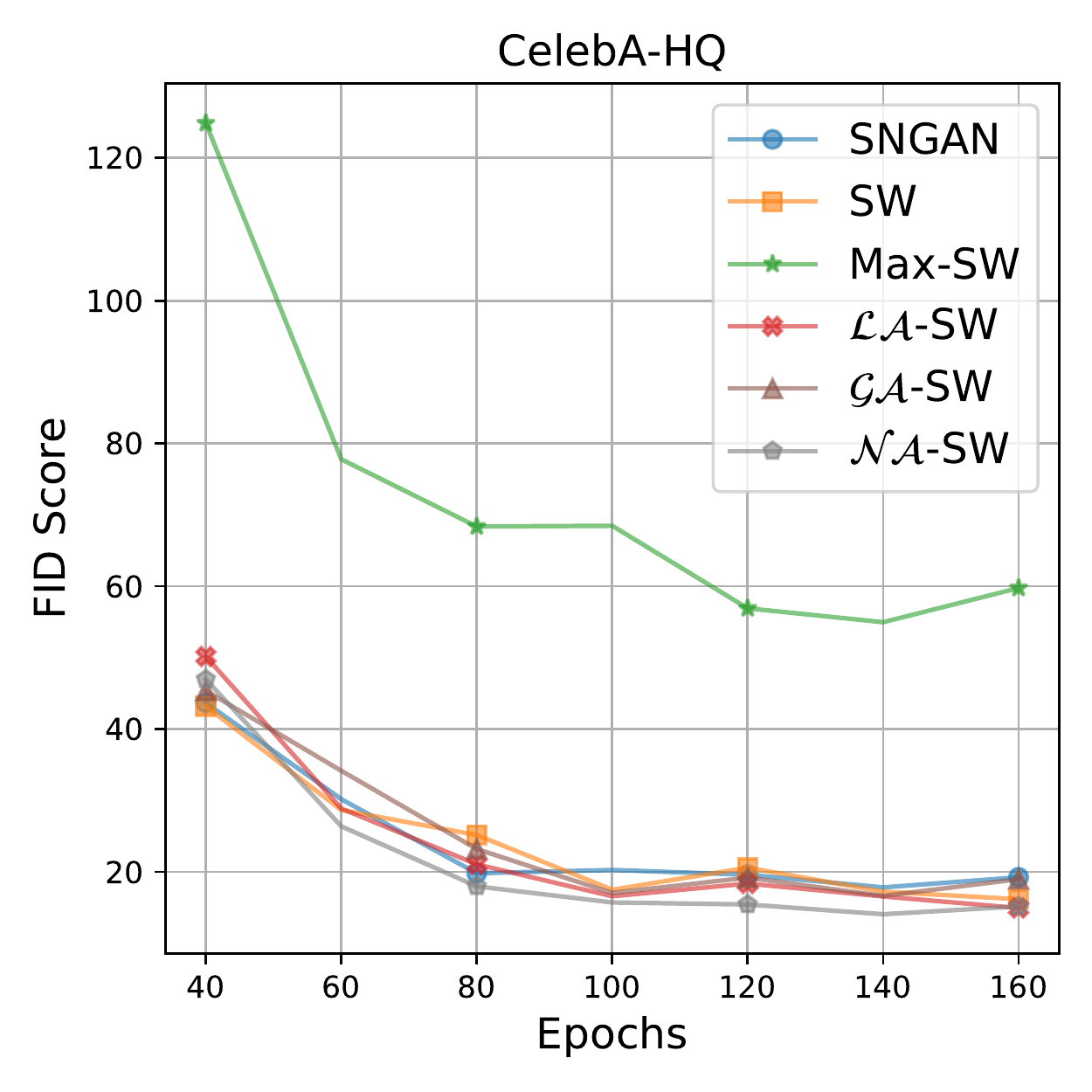} 
  &
\widgraph{0.23\textwidth}{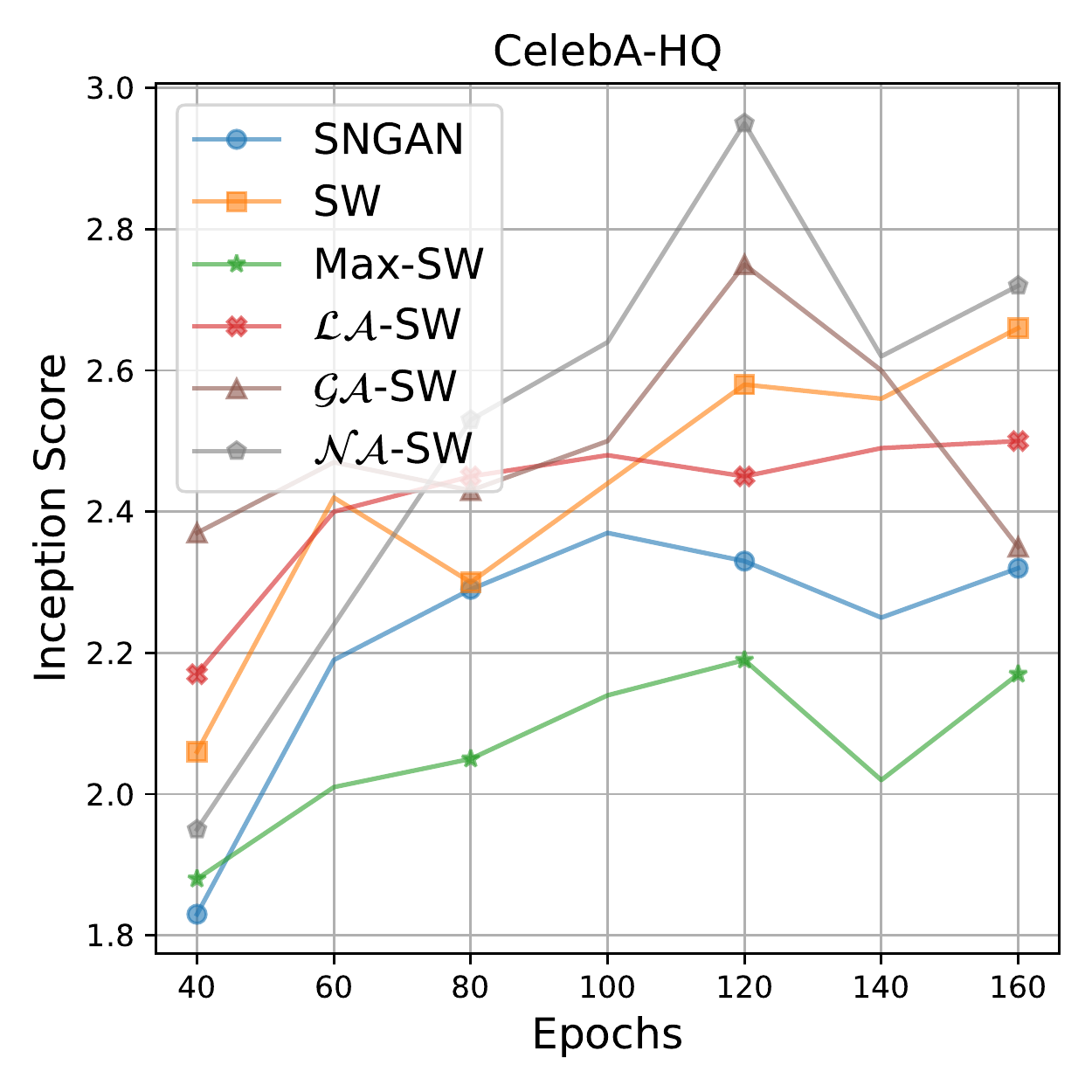} 
  \end{tabular}
  \end{center}
  \vskip -0.2in
  \caption{
  \footnotesize{FID scores and IS scores over epochs of different training losses on datasets. We observe that members of $\ASW$ usually help the generative models converge faster.
}
} 
  \label{fig:iterCIFAR}
\end{figure*}
We first demonstrate the quality of using $\ASW$ in the training generative model compared to the baseline SNGAN, and other mini-batch sliced Wasserstein variants. Then, we investigate the convergence of generative models trained by different losses including the standard SNGAN's loss, mini-batch SW, mini-batch Max-SW, and $\ASW$ by looking at their FID scores and IS scores over training epochs of their best settings. After that, we compare models qualitatively by showing their randomly generated images. Finally, we report the training speed (number of training iterations per second) and the training memory (megabytes) of all settings of all training losses.

\begin{table}[!t]
    \centering
    \caption{\footnotesize{Summary of FID and IS scores of methods on CIFAR10 (32x32), CelebA (64x64), STL10 (96x96), and CelebA-HQ (128x128). We observe that $\ASW$ losses provide the best results among all the training losses.}}
    \scalebox{0.9}{
    \begin{tabular}{l|cc|cc|cc|cc}
    \toprule
     \multirow{2}{*}{Method}& \multicolumn{2}{c|}{CIFAR10 (32x32)}&\multicolumn{2}{c|}{CelebA (64x64)}&\multicolumn{2}{c|}{STL10 (96x96)}&\multicolumn{2}{c}{CelebA-HQ (128x128)}\\
     \cmidrule{2-9}
     & FID ($\downarrow$) &IS ($\uparrow$)& FID ($\downarrow$) &IS ($\uparrow$)& FID ($\downarrow$) &IS ($\uparrow$)& FID ($\downarrow$) &IS ($\uparrow$)\\
    \midrule
         SNGAN & 17.09 & 8.07&12.41 &2.61 &59.48&9.29 &19.25&2.32\\
         SW&14.25$\pm$0.84&8.12$\pm$0.07&10.45&2.70&56.32&10.37&16.17&2.65\\
         Max-SW& 31.33$\pm$3.02& 6.67$\pm$0.37&11.28&2.60&77.40&9.46&29.50&2.36\\
        \midrule
         $\LASW$ (ours) & \textbf{13.21$\pm$0.69 }&8.19$\pm$0.03&9.82&2.72&\textbf{52.08}&\textbf{10.52}&\textbf{14.94}&2.50\\
          $\GASW$ (ours) & 13.64$\pm$0.11& 8.22$\pm$0.11 &9.21&2.78&53.80&10.40&18.97&2.34\\
         $\NASW$  (ours) & 14.22$\pm$0.51&\textbf{8.29$\pm$0.08}&\textbf{8.91}&\textbf{2.82} &53.90&10.14&15.17&\textbf{2.72}\\
         \bottomrule
    \end{tabular}
    }
    \label{tab:summary}
    \vspace{-0.5 em}
\end{table}

\textbf{Summary of FID and IS scores:} We show FID scores and IS scores of all models at the last training step on all datasets in Table~\ref{tab:summary}. For SW and Max-SW, we select the best setting of hyperparameters for each score. In particular, we search for the best setting of the number of projections $L \in \{1,100,1000,10000\}$. Also, we do a grid search on two hyperparameters of Max-SW, namely, the slice maximum number of iterations $T_2 \in \{1,10,100\}$ and the slice learning rate $\eta_2 \in \{0.001,0.01,0.1\}$. The detailed FID scores and IS scores for all settings are reported in Table~\ref{tab:detailFID} in Appendix~\ref{sec:fullexp}. For amortized models, we fix the slice learning rate $\eta_2 =0.01$.  From Table~\ref{tab:summary}, the best amortized model provides lower FID scores and IS scores than SNGAN, SW, and Max-SW on all datasets of multiple image resolutions. We would like to recall that, SNGAN is reported to be better than WGAN~\cite{arjovsky2017wasserstein} in~\cite{miyato2018spectral}. Furthermore, the best generative models trained by $\ASW$ are better than models trained with SNGAN, SW, and Max-SW. Interestingly, the $\LASW$ performs consistently well compared to other members of $\ASW$. Also, we observe that Max-SW performs worse than both $\ASW$ and SW. This might be because the local optimization of Max-SW gets stuck at some bad optimum. However, we would like to recall that Max-SW is still better than SW with $L=1$ (see Table~\ref{tab:detailFID} in Appendix~\ref{sec:fullexp}). It emphasizes the benefit of searching for a good direction for projecting.

\textbf{FID and IS scores over training epochs:} We show the values of FID scores and Inception scores over epochs on CIFAR10, CelebA, STL10, and CelebA-HQ in Figure~\ref{fig:iterCIFAR}. According to the figures in Figure~\ref{fig:iterCIFAR}, we observe that using SW and $\ASW$ helps the generative models converge faster than SNGAN. Moreover, FID lines of $\ASW$ are usually under the lines of other losses and the IS lines of $\ASW$ are usually above the lines of others. Therefore, $\ASW$ losses including $\LASW$, $\GASW$, and $\NASW$ can improve the convergence of training generative models.

\begin{figure*}[!t]
\begin{center}
    
  \begin{tabular}{ccc}
  \widgraph{0.28\textwidth}{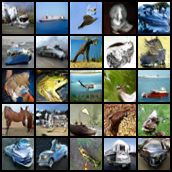} 
  &
\widgraph{0.28\textwidth}{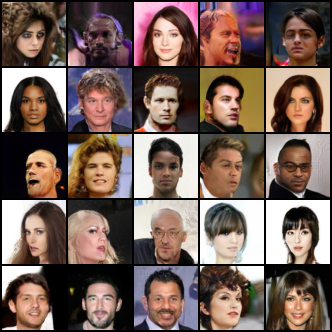} 
&
\widgraph{0.28\textwidth}{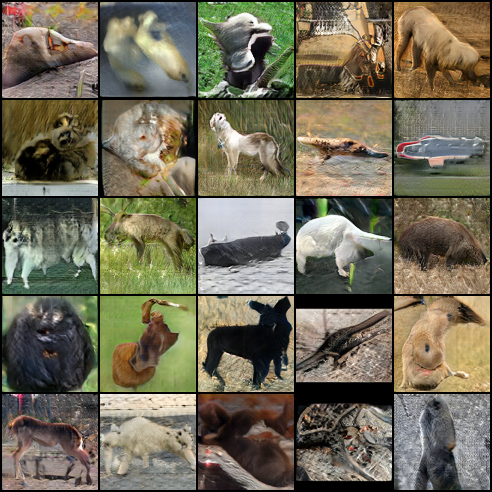} 
\\
SNGAN (CIFAR) & SNGAN (CelebA) & SNGAN (STL10) 
\\
  \widgraph{0.28\textwidth}{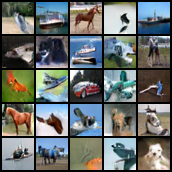} 
  &
\widgraph{0.28\textwidth}{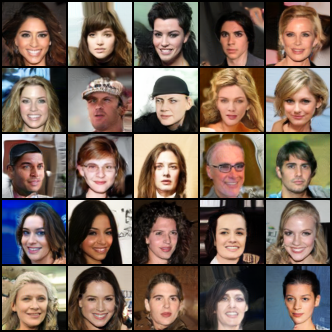} 
&
\widgraph{0.28\textwidth}{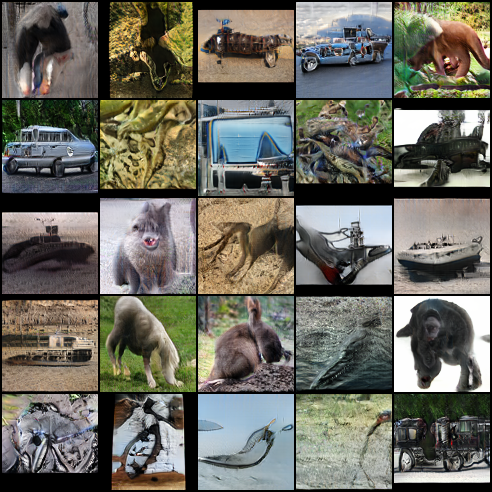} 
\\
$\LASW$ (CIFAR) & $\LASW$  (CelebA) & $\LASW$  (STL10)
  \end{tabular}
  \end{center}
  \vskip -0.05in
  \caption{
  \footnotesize{Random generated images of SNGAN and $\LASW$ from CIFAR10, CelebA, and  STL10.
}
} 
  \label{fig:lasw1}
\end{figure*}

\textbf{Generated images:} We show generated images on CIFAR10, CelebA, STL10 from SNGAN, and $\LASW$ in Figure~\ref{fig:lasw1} as a qualitative comparison. The generated images on CelebAHQ and the generated images of Max-SW, $\GASW$, and $\NASW$ are given in Appendix~\ref{sec:fullexp}. From these images, we observe that the quality of generated images is consistent with the FID scores and the IS scores. Therefore, it reinforces the benefits of using $\ASW$ to train generative models. Again, we would like to recall that all generated images are completely random without cherry-picking.

\begin{table}[!t]
    \centering
    \caption{\footnotesize{Computational time and memory of methods (in iterations per a second and megabytes (MB)).}}
    \scalebox{0.8}{
    \begin{tabular}{l|cc|cc|cc|cc}
    \toprule
     \multirow{2}{*}{Method}& \multicolumn{2}{c|}{CIFAR10 (32x32)}&\multicolumn{2}{c|}{CelebA (64x64)}&\multicolumn{2}{c|}{STL10 (96x96)}&\multicolumn{2}{c}{CelebA-HQ} (128x128)\\
     \cmidrule{2-9}
     & Iters/s ($\uparrow$) &Mem ($\downarrow$)& Iters/s ($\uparrow$) &Mem ($\downarrow$)& Iters/s ($\uparrow$) &Mem ($\downarrow$)& Iters/s ($\uparrow$) &Mem ($\downarrow$)\\
    \midrule
         SNGAN (baseline)& 19.97& 1740&6.31 &6713 &9.33&3866& 10.41 &3459\\
         \midrule
         SW (L=1)& 18.73&2078&6.17 &8011&9.31&4597&10.25&4111 \\ 
         SW (L=100)& 18.42&2093&6.15&8015&9.11&4609&10.17&4120\\ 
         SW (L=1000)& 14.96&2112 &6.13 &8047&9.03&4616&9.63&4143 \\ 
         SW (L=10000)& 5.84&2421 &4.21&8353&6.50&4780&5.17&4428\\ 
         \midrule
          Max-SW ($T_2$=1) & 18.61 & 2078 &6.17&8011& 9.23& 4597 &10.22 &4111\\
         Max-SW ($T_2$=10) & 18.16 & 2078 &6.15&8011&9.17 &4597 &10.16 &4111\\
         Max-SW ($T_2$=100) & 13.47& 2078 &5.78&8011 &8.32 &4597&8.13 &4111\\
         \midrule
         $\LASW$ (ours) & 18.58&2086&6.17&8021&9.23&4600  & 10.19& 4115\\
          $\GASW$ (ours) & 17.27& 4151 &6.07&10083&9.08&5251 & 10.11&6163\\
         $\NASW$  (ours) & 17.67&4134&6.13&10068 &9.11&5249&10.15 &6152\\
         \bottomrule
    \end{tabular}
    }
    \label{tab:timeandmem}
\end{table}

\textbf{Computational time and memory:} We report the number of training iterations per second and the memory in megabytes (MB) in Table~\ref{tab:timeandmem}. We would like to recall that reported numbers are under some errors due to the state of the computational device. From the table, we see that $\LASW$ is comparable to Max-SW and SW $(L=1)$ about the computational memory and the computational time. More importantly, $\LASW$ is faster and consumes less memory than SW ($L \geq 100$) and Max-SW ($T_2 \geq 10$). Compared to SNGAN, SW variants increase the demand for memory and computation slightly. From $\LASW$ to $\GASW$ and $\NASW$, the computational time is slower slightly; however, we need between 800 to 2100 MB of memory in extra. Again, the additional memory depends on the chosen parameterization (see Section~\ref{sec:ASW}). From this table, we can see that using sliced Wasserstein models gives better generative quality than SNGAN but it also costs more computational time and memory. 
Among sliced Wasserstein variants, $\LASW$ is the best option since it costs the least additional memory and time while it gives consistently good results. We refer to Section~\ref{sec:ASW} for discussion of the time and projection complexities of $\ASW$. 

\section{Conclusion}
\label{sec:conclusion}

We propose using amortized optimization for speeding up the training of generative models that are based on mini-batch sliced Wasserstein with projection optimization. We introduce three types of amortized models, including the linear, generalized, and non-linear amortized models, for predicting optimal projecting directions between all pairs of mini-batch probability measures. Moreover, using three types of amortized models leads to three corresponding mini-batch losses which are the linear amortized sliced Wasserstein, the generalized linear amortized sliced Wasserstein, and  the non-linear amortized sliced Wasserstein. We then show that these losses can improve the result of training deep generative models in both training speed and generative performance. 

\section*{Acknowledgements}
NH
acknowledges support from the NSF IFML 2019844 and the NSF AI Institute for Foundations of Machine Learning.
\clearpage
\bibliography{example_paper}
\bibliographystyle{abbrv}
\section*{Checklist}

\begin{enumerate}
\item For all authors... 
\begin{enumerate}
\item Do the main claims made in the abstract and introduction accurately
reflect the paper's contributions and scope? \answerYes{} 
\item Did you describe the limitations of your work? \answerYes{} 
\item Did you discuss any potential negative societal impacts of your work?
\answerYes{} 
\item Have you read the ethics review guidelines and ensured that your paper
conforms to them? \answerYes{} 
\end{enumerate}
\item If you are including theoretical results... 
\begin{enumerate}
\item Did you state the full set of assumptions of all theoretical results?
\answerYes{} 
\item Did you include complete proofs of all theoretical results? \answerYes{} 
\end{enumerate}
\item If you ran experiments... 
\begin{enumerate}
\item Did you include the code, data, and instructions needed to reproduce
the main experimental results (either in the supplemental material
or as a URL)? \answerYes{} 
\item Did you specify all the training details (e.g., data splits, hyperparameters,
how they were chosen)? \answerYes{} 
\item Did you report error bars (e.g., with respect to the random seed after
running experiments multiple times)? \answerYes{} 
\item Did you include the total amount of compute and the type of resources
used (e.g., type of GPUs, internal cluster, or cloud provider)? \answerYes{} 
\end{enumerate}
\item If you are using existing assets (e.g., code, data, models) or curating/releasing
new assets... 
\begin{enumerate}
\item If your work uses existing assets, did you cite the creators? \answerYes{} 
\item Did you mention the license of the assets? \answerNA{} 
\item Did you include any new assets either in the supplemental material
or as a URL? \answerYes{} 
\item Did you discuss whether and how consent was obtained from people whose
data you're using/curating? \answerNA{} 
\item Did you discuss whether the data you are using/curating contains personally
identifiable information or offensive content? \answerNA{} 
\end{enumerate}
\item If you used crowdsourcing or conducted research with human subjects... 
\begin{enumerate}
\item Did you include the full text of instructions given to participants
and screenshots, if applicable? \answerNA{} 
\item Did you describe any potential participant risks, with links to Institutional
Review Board (IRB) approvals, if applicable? \answerNA{} 
\item Did you include the estimated hourly wage paid to participants and
the total amount spent on participant compensation? \answerNA{} 
\end{enumerate}
\end{enumerate}

\newpage
\appendix
\begin{center}
{\bf{\large{Supplement to "Amortized Projection Optimization for Sliced Wasserstein Generative Models"}}}
\end{center}

In this supplement, we first collect some proofs in Appendix~\ref{sec:proof}. We then introduce Amortized Projected Robust Wasserstein in Appendix~\ref{sec:APRW}. Next, we discuss the training detail of generative models with different mini-batch losses in Appendix~\ref{sec:training_detail}. Moreover, we present detailed results on the deep generative model in Appendix~\ref{sec:fullexp}. Next, we report the experimental settings including neural network architectures, and hyper-parameter choices in Appendix~\ref{sec:settings}. Finally, we discuss the potential impacts of our works in Appendix~\ref{sec:potential}.
\section{Proofs}
\label{sec:proof}
In this appendix, we provide proofs for main results in the main text.
\subsection{Proof of Proposition~\ref{proposition:amortized_sliced}}
\label{subsec:proof:proposition:amortized_sliced}
Recall that, the definition of $\ASW(\mu,\nu)$ is as follows:
\begin{align*}
    \ASW(\mu,\nu) = \max_{\psi \in \Psi}\mathbb{E}_{(X,Y) \sim \mu^{\otimes m} \otimes \nu^{\otimes m}}[\text{W}_p(f_\psi (X,Y) \sharp P_X,f_\psi (X,Y) \sharp P_Y) ].
\end{align*}
For the symmetric property of the amortized sliced Wasserstein, we have
\begin{align*}
    \ASW(\nu,\mu) 
    &= \max_{\psi \in \Psi}\mathbb{E}_{(Y,X) \sim \nu^{\otimes m} \otimes \mu^{\otimes m}}[\text{W}_p(f_\psi (Y,X) \sharp P_X,f_\psi (Y,X) \sharp P_Y ] \\
    &= \max_{\psi \in \Psi} \mathbb{E}_{(Y,X) \sim \nu^{\otimes m} \otimes \mu^{\otimes m}}[\text{W}_p(f_\psi (X,Y) \sharp P_X,f_\psi (X,Y) \sharp P_Y ] \\
    &= \max_{\psi \in \Psi}  \mathbb{E}_{(X,Y) \sim \mu^{\otimes m} \otimes \nu^{\otimes m}}[\text{W}_p(f_\psi (X,Y) \sharp P_X,f_\psi (X,Y) \sharp P_Y) ] \\
    &= \ASW (\mu,\nu),
\end{align*}
where the second equality is because of the symmetry of Wasserstein distance, the third equality is due to the symmetry of $f_\psi(X,Y)$ (see forms of $f_\psi(X,Y)$ in Section~\ref{sec:ASW}). The positiveness of $\ASW$ comes directly from the non-negativity of the Wasserstein distance.

To prove that $\ASW$ violates the identity, we use a counter example  where $\mu =\nu= \frac{1}{2} \delta_{x_1} + \frac{1}{2} \delta_{x_2} $ ($x_1\neq x_2$). In this example, there exists a pair of mini-batches $X=(x_1,x_1)$ and $Y=(x_2,x_2)$. We choose $f_\psi (X,Y)=\frac{x_1+x_2}{||x_1+x_2||_2}$, then $f_\psi (X,Y) \sharp P_X \neq f_\psi (X,Y) \sharp P_Y$ which implies $\text{W}_p(f_\psi (X,Y) \sharp P_X,f_\psi (X,Y) \sharp P_Y)>0$. Since $\ASW$ defines on the maximum value of $\psi \in \Psi$, $\ASW(\mu,\nu) \geq \text{W}_p(f_\psi (X,Y) \sharp P_X,f_\psi (X,Y) \sharp P_Y)>0 $.
\subsection{Proof of Proposition~\ref{prop:lower_bound}}
\label{subsec:proof:lower_bound}
Since the function $f_{\psi}$ is continuous in terms of $\psi$, it indicates that the function $\mathbb{E}_{(X,Y) \sim \mu^{\otimes m} \otimes \nu^{\otimes m}}[\text{W}_p(f_\psi (X,Y) \sharp P_X,f_\psi (X,Y) \sharp P_Y)]$ is continuous in terms of $\psi$. Furthermore, as the parameter space $\Psi$ is compact, there exist $\psi^* \in \argmax_{\psi \in \Psi}\mathbb{E}_{(X,Y) \sim \mu^{\otimes m} \otimes \nu^{\otimes m}}[\text{W}_p(f_\psi (X,Y) \sharp P_X,f_\psi (X,Y) \sharp P_Y) ]$. Then, we have
\begin{align*}
    \ASW(\mu,\nu) &= \mathbb{E}_{(X,Y) \sim \mu^{\otimes m} \otimes \nu^{\otimes m}}[\text{W}_p(f_{\psi^*} (X,Y) \sharp P_X,f_{\psi^*} (X,Y) \sharp P_Y) ] \\
    &=  \mathbb{E}_{(X,Y) \sim \mu^{\otimes m} \otimes \nu^{\otimes m}}[\text{W}_p(\theta_{\psi^\star} \sharp P_X,\theta_{\psi^\star}  \sharp P_Y) ] \\
    &\leq \mathbb{E}_{(X,Y) \sim \mu^{\otimes m} \otimes \nu^{\otimes m}}\left[\max_{\theta \in \mathbb{S}^{d-1}}\text{W}_p(\theta \sharp P_X,\theta  \sharp P_Y) \right] := \text{m-Max-SW}(\mu,\nu).
\end{align*}
As a consequence, we obtain the conclusion of the proposition.
\begin{table}[]
    \centering
    \caption{\footnotesize{Summary of FID and IS scores of methods on CIFAR10 (32x32), CelebA (64x64), STL10 (96x96), and CelebA-HQ (128x128).}}
    \scalebox{0.8}{
    \begin{tabular}{l|cc|cc|cc|cc}
    \toprule
     \multirow{2}{*}{Method}& \multicolumn{2}{c|}{CIFAR10 (32x32)}&\multicolumn{2}{c|}{CelebA (64x64)}&\multicolumn{2}{c|}{STL10 (96x96)}&\multicolumn{2}{c|}{CelebA-HQ (128x128)}\\
     \cmidrule{2-9}
     & FID ($\downarrow$) &IS ($\uparrow$)& FID ($\downarrow$) &IS ($\uparrow$)& FID ($\downarrow$) &IS ($\uparrow$)& FID ($\downarrow$) &IS ($\uparrow$)\\
    \midrule
         SNGAN (baseline)& 17.09 & 8.07&12.41 &2.61 &59.48&9.29 &19.25&2.32\\
         \midrule
         SW (L=1)& 53.95&5.41&34.47 &2.61&144.64&5.82 &147.35&2.02\\ 
         SW (L=100)& 15.90$\pm$0.45 &8.08$\pm$0.04 &10.45&2.70&62.44&9.91 &17.57&2.43\\ 
         SW (L=1000)& 14.58$\pm$0.95&8.10$\pm$0.06 &10.96 &2.67&57.12&10.25&16.17&2.65 \\ 
         SW (L=10000)& 14.25$\pm$0.84&8.12$\pm$0.07&10.82&2.66&56.32&10.37&18.08&2.62\\ 
         \midrule
          Max-SW ($T_2$=1; $\eta_2$=0.001) & 35.52$\pm$1.97 & 6.54$\pm$0.22 &11.28&2.60 &101.37 &7.98  & 34.97&1.98\\
         Max-SW ($T_2$=10;$\eta_2$=0.001) & 31.33$\pm$3.02 & 6.67$\pm$0.37 &15.98&2.51 & 77.40&9.46  & 29.50&2.36\\
         Max-SW ($T_2$=100; $\eta_2$=0.001) & 41.20$\pm$2.33& 6.02$\pm$0.25&16.52&2.46 &86.91 & 9.05& 56.20&2.26\\
         Max-SW ($T_2$=1; $\eta_2$=0.01) & 40.28$\pm$2.10 & 6.21$\pm$0.19 &14.11&2.62 &88.29 & 9.26& 43.16&2.36\\
         Max-SW ($T_2$=10; $\eta_2$=0.01) & 39.56$\pm$4.55 & 6.25$\pm$0.36 &16.89 &2.49 &90.82 &9.18 & 59.74&2.16\\
         Max-SW ($T_2$=100; $\eta_2$=0.01) & 44.68$\pm$3.22& 5.98$\pm$0.31&12.80&2.70 &99.32 &8.52 & 55.94&2.11\\ 
          Max-SW ($T_2$=1; $\eta_2$=0.1) & 36.60& 6.58&18.87&2.42 & 94.33& 8.19&  52.68&2.16\\ 
           Max-SW ($T_2$=10; $\eta_2$=0.1) & 48.42& 6.19&16.22&2.49 &90.17 &9.70 & 43.65&2.17\\ 
           Max-SW ($T_2$=100; $\eta_2$=0.1) & 50.74& 5.42&14.40&2.59 & 101.38& 8.46 &42.81&2.20\\ 
           \midrule
         $\LASW$ (ours) & \textbf{13.21$\pm$0.69 }&8.19$\pm$0.03&9.82&2.72&\textbf{52.08}&\textbf{10.52}&\textbf{14.94}&2.50\\
          $\GASW$ (ours) & 13.64$\pm$0.11& 8.22$\pm$0.11  &9.21&2.78&53.80&10.40&18.97&2.34\\
         $\NASW$  (ours) & 14.22$\pm$0.51&\textbf{8.29$\pm$0.08}&\textbf{8.91}&\textbf{2.82} &53.90&10.14&15.17&\textbf{2.72}\\
         \bottomrule
    \end{tabular}
    }
    \label{tab:detailFID}
\end{table}


\section{Amortized Projected Robust Wasserstein}
\label{sec:APRW}

We first recall the definition of projected robust Wasserstein (PRW) distance~\cite{paty2019subspace}. Given two probability measures $\mu,\nu \in \mathcal{P}_p(\mathbb{R}^d)$, the projected robust Wasserstein distance between $\mu$ and $\nu$ is defined as:
\begin{align}
    PRW_k(\mu,\nu):= \max_{U \in \mathbb{V}_k(\mathbb{R}^{d})} W_p(U\sharp \mu,U\sharp \nu),
\end{align}
where $\mathbb{V}_k(\mathbb{R}^d):=\{ U \in \mathbb{R}^{d\times k}|U^\top U = I_k\}$ is the Stefel Manifold. PRW can be seen as the generalization of Max-SW since PRW with $k=1$ is equivalent to Max-SW. Similar to Max-SW, the optimization of PRW is solved by using projected gradient ascent. The detailed of the algorithm is given in Algorithm~\ref{alg:PRW}. We would like to recall that other methods of optimization have also been used to solved PRW such as Riemannian optimization~\cite{lin2020projection}, block coordinate descent~\cite{pmlr-v139-huang21e}. However, in this paper, we consider the original and simplest method which is projected gradient ascent.

In deep learning and large-scale applications, the mini-batch loss version of PRW is used, that is defined as follow:
\begin{align}
    \text{m-}PRW_k(\mu, \nu)=\mathbb{E}_{X, Y \sim \mu^{\otimes m} \otimes \nu^{\otimes m}}\left[\max_{U \in \mathbb{V}_k(\mathbb{R}^d)} W_p(U\sharp P_X,U\sharp P_Y)\right].
\end{align}

\textbf{Amortized Projected Robust Wasserstein loss:} We  define Amortized Projected Rubust Wasserstein loss as follow:
\begin{definition}
\label{def:APRW}
Let $p\geq 1$, $m\geq 1$, and $\mu,\nu$ are two probability measures in $\mathcal{P}(\mathbb{R}^d)$. Given an amortized model $f_\psi: \mathbb{R}^{dm}\times \mathbb{R}^{dm} \to \mathbb{V}_k(\mathbb{R}^d)$ where $\psi \in \Psi$, the amortized projected robust Wasserstein between $\mu$ and $\nu$ is:
\begin{align}
    \mathcal{A}\text{-}PRW(\mu,\nu) :=\max_{\psi \in \Psi} \mathbb{E}_{(X,Y) \sim \mu^{\otimes m} \otimes \nu^{\otimes m}} [W_{p}\left(f_{\psi}(X, Y) \sharp P_{X}, f_{\psi}(X, Y) \sharp P_{Y}\right)].
\end{align}
\end{definition}
Similar to the case of $\ASW$, $\mathcal{A}$-PRW is symmetric, positive, and is a lowerbound of PRW. Also, $\mathcal{A}$-PRW  is not a metric since it does not satisfy the identity property.

\textbf{Amortized models:} Similar to the case of $\ASW$, we can derive linear model, generalized linear model, and non-linear amortized model. The only change is that the model gives $k$ output vectors instead of $1$ vector. 
\begin{definition}
\label{def:linearPRWmodel}
Given $X,Y \in \mathbb{R}^{dm}$, and the one-one "reshape" maping $T: \mathbb{R}^{dm} \to \mathbb{R}^{d\times m}$, the linear projected amortized model is defined as:
\begin{align}
    f_\psi (X,Y) := \text{Proj}_{\mathbb{V}_k(\mathbb{R}^d)}(W_{0}+T(X) W_{1}+T(Y) W_{2}),
\end{align}
where $W_1,W_2 \in \mathbb{R}^{m \times k}, W_0 \in \mathbb{R}^{d\times k}$, and $\text{Proj}_{\mathbb{V}_k(\mathbb{R}^d)}$ return the $Q$ matrix in QR decomposition.
\end{definition}
The definitions of the generalized linear projected amortized model and non-linear projected amortized model are straight-forward from the definitions of generalized linear model and non-linear model in $\ASW$.

\begin{algorithm}[!t]
\caption{Projected Robust Wasserstein distance}
\begin{algorithmic}
\label{alg:PRW}
\STATE \textbf{Input:} Probability measures: $\mu,\nu$, learning rate $\eta$, max number of iterations $T$.
  \STATE Initialize $U$
  \WHILE{$U$ not converge or reach $T$}
  \STATE $U = U +  \eta \cdot \nabla_U\text{W}_p (U \sharp \mu,U \sharp \nu)$
  \STATE $Q,R=QR(U)$ (QR decomposition)
  \STATE $U=Q$
  \ENDWHILE
 \STATE \textbf{Return:} $\theta$
\end{algorithmic}
\end{algorithm}

\section{Training Generative Models}
\label{sec:training_detail}
In this section, we review the parameterization of training losses of generative models.
 
\textbf{Parametrization:} We first discuss the parametrization of the model distribution $\nu_\phi$. In particular, $\nu_\phi$ is a pushforward probability measure that is created by pushing a unit multivariate Gaussian ($\epsilon$) through a neural network $G_\phi$ that maps from the realization of the noise to the data space. The detail of the architecture of $G_\phi$ is given in Appendix~\ref{sec:settings}. For training both SNGAN and generative models of SW, Max-SW, and $\ASW$, we need a second neural network $T_\beta$ that maps from data space to a single scalar. The second neural network is called \textit{Discriminator} in SNGAN or \textit{Feature encoder} in the others. However, the architecture of the second neural network is the same for all models (see Appendix~\ref{sec:settings}). For the better distinction between training objectives of SNGAN and the objectives of the others, we denote $T_{\beta_1}$ is the sub neural network of $T_\beta$ that maps from the data space to a feature space (output of the last Resnet block), and $T_{\beta_2}$ that maps from the feature space (image of $T_{\beta_1}$) to a single scalar. More precisely, $T_\beta = T_{\beta_2} \circ T_{\beta_1}$. Again, we specify $T_{\beta_1}$ and $T_{\beta_1}$ in Appendix~\ref{sec:settings}.
\begin{figure*}[!t]
\begin{center}
  \begin{tabular}{ccc}
  \widgraph{0.3\textwidth}{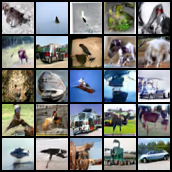} 
  &
\widgraph{0.3\textwidth}{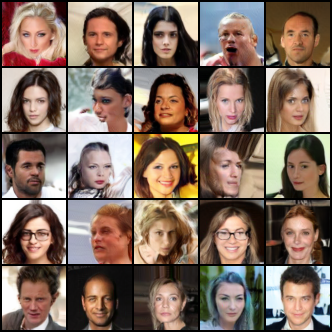} 
&
\widgraph{0.3\textwidth}{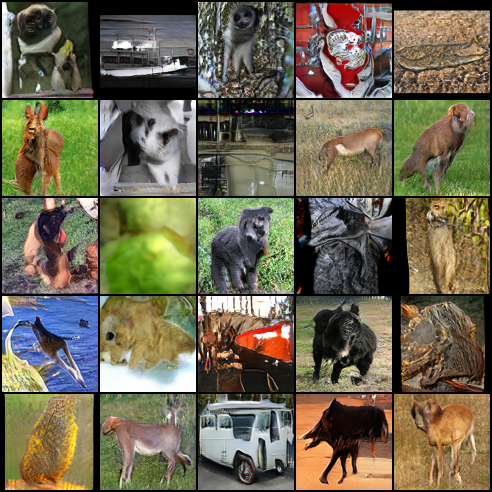} 
\\
SW (CIFAR) & SW (CelebA) & SW (STL10) 
\\
\widgraph{0.3\textwidth}{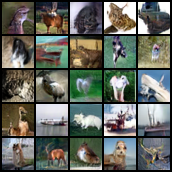} 
  &
\widgraph{0.3\textwidth}{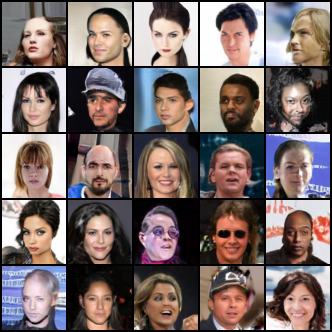} 
&
\widgraph{0.3\textwidth}{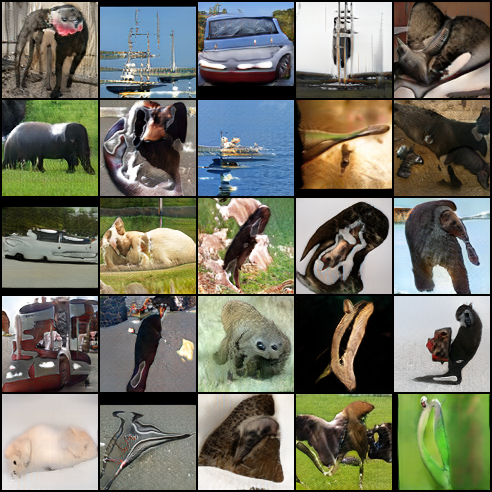} 
\\
$\GASW$ (CIFAR) & $\GASW$ (CelebA) & $\GASW$ (STL10) 
\\
\widgraph{0.3\textwidth}{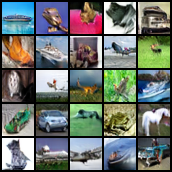} 
  &
\widgraph{0.3\textwidth}{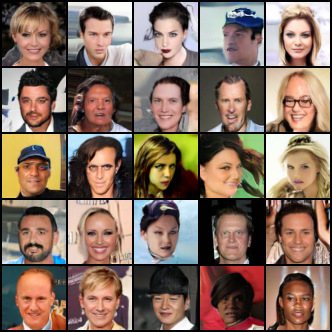} 
&
\widgraph{0.3\textwidth}{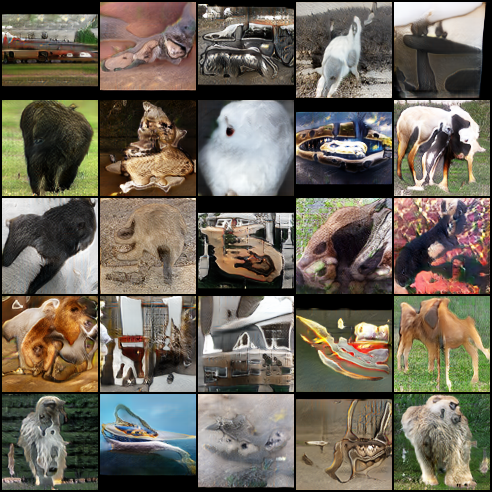} 
\\
$\NASW$ (CIFAR) & $\NASW$ (CelebA) & $\NASW$ (STL10) 
  \end{tabular}
  \end{center}
  \vskip -0.1in
  \caption{
  \footnotesize{Random generated images of SW, $\GASW$, and $\NASW$ from CIFAR10, CelebA, and STL10.
}
} 
  \label{fig:sngan1}
  \vskip -0.1in
\end{figure*}

\textbf{Training SNGAN:} Let $\mu$ is theta data probability measure,  these two optimization problems are done alternatively in training SNGAN:
\begin{align*}
    &\min_{\beta_1,\beta_2} \left(\mathbb{E}_{x \sim \mu} [\min (0,-1+ T_{\beta_2}(T_{\beta_1}(x)))] + \mathbb{E}_{z \sim \epsilon} [\min(0, -1- T_{\beta_2}(T_{\beta_1} (G_\phi(z))))] \right), \\
    &\min_{\phi} \mathbb{E}_{z \sim \epsilon}[-T_{\beta_2}(T_{\beta_1} (G_\phi(z))) ].
\end{align*}

\textbf{Training SW, Max-SW, and $\ASW$:} For training these models, we adapt the framework in~\cite{deshpande2018generative} to SNGAN, namely, we use these two objectives:
\begin{align*}
    &\min_{\beta_1,\beta_2} \left(\mathbb{E}_{x \sim \mu} [\min (0,-1+ T_{\beta_2}(T_{\beta_1}(x)))] + \mathbb{E}_{z \sim \epsilon} [\min(0, -1-T_{\beta_2}(T_{\beta_1} (G_\phi(z))))] \right), \\
    &\min_{\phi} \tilde{\mathcal{D}}(\tilde{T}_{\beta_1,\beta_2} \sharp \mu, \tilde{T}_{\beta_1,\beta_2}\sharp G_\phi \sharp \epsilon),
\end{align*}
where the function $\tilde{T}_{\beta_1,\beta_2} = [T_{\beta_1}(x), T_{\beta_2}(T_{\beta_1}(x))]$ which is the concatenation vector of $T_{\beta_1}(x)$ and $T_{\beta_2}(T_{\beta_1}(x))$, $\mathcal{D}$ is one of the mini-batch SW,  the mini-batch Max-SW (see Equation~\ref{eq:mmaxsw}), and $\ASW$ (see Definition~\ref{def:asw}). This technique is an application of metric learning since $\mathcal{L}_p$ norm is not meaningful on the space of natural images. This observation is mentioned in previous works~\cite{deshpande2018generative,genevay2018learning,stanczuk2021wasserstein,nguyen2021distributional}.

\textbf{Other settings:} The information about the mini-batch size, the learning rate, the optimizer, the number of iterations, and so on, are given in Appendix~\ref{sec:settings}.
\section{Full Experimental Results}
\label{sec:fullexp}

\textbf{Detailed FID scores and Inception scores:}  We first show the detailed FID scores and IS scores of all settings in Table~\ref{tab:detailFID}. From the table, we can see that the quality of the SW depends on the number of projections. Namely, a higher number of projections often leads to better performance. For Max-SW, we obverse that increasing the number of iterations $T_2$ might not lead to a lower FID score and a higher IS score. The reason might be that the optimization gets stuck at some local optima. For the choice of the learning rate $\eta_2$, we do not see any superior setting for Max-SW.

\begin{figure*}[!t]
\begin{center}
  \begin{tabular}{ccc}
\widgraph{0.3\textwidth}{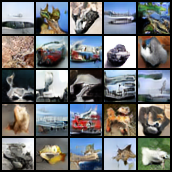} 
  &
\widgraph{0.3\textwidth}{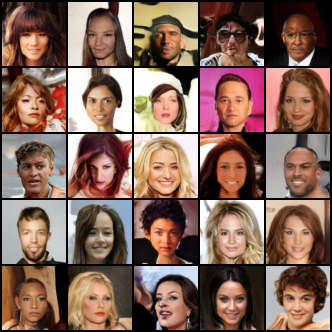} 
&
\widgraph{0.3\textwidth}{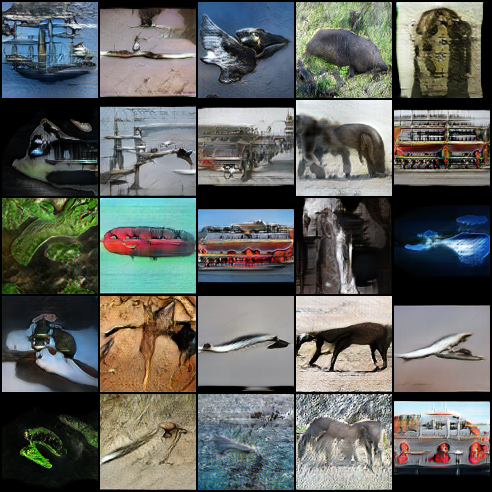} 
\\
Max-SW (CIFAR) & Max-SW  (CelebA) & Max-SW  (STL10) 
  \end{tabular}
  \end{center}
  \vskip -0.1in
  \caption{
  \footnotesize{Random generated images of Max-SW from CIFAR10, CelebA, and STL10.
}
} 
  \label{fig:sw1}
\end{figure*}

\begin{figure*}[!t]
\begin{center}
    
  \begin{tabular}{cc}
  \widgraph{0.45\textwidth}{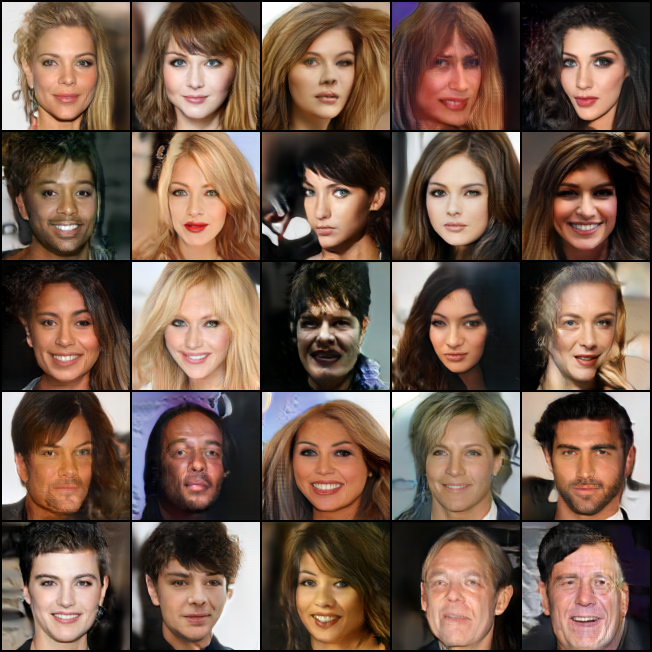} 
  &
\widgraph{0.45\textwidth}{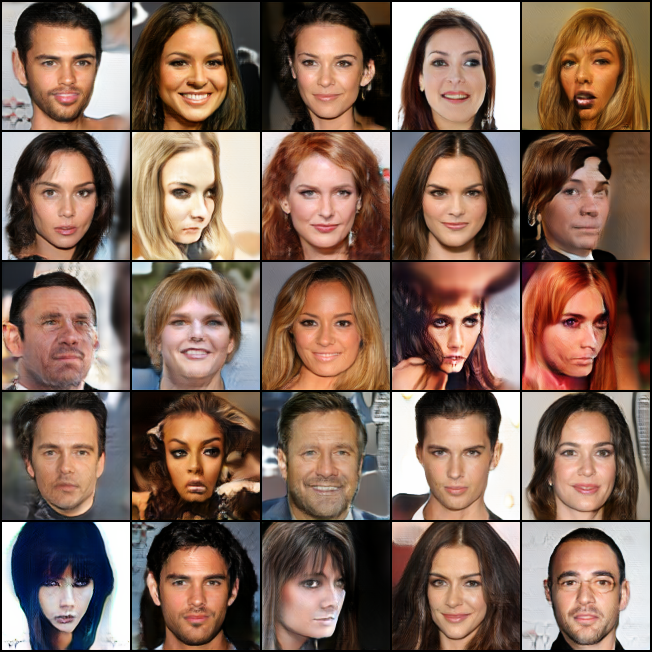} 
\\
SNGAN (CelebA-HQ)  &$\LASW$ (CelebA-HQ)
  \end{tabular}
  \end{center}
  \vskip -0.1in
  \caption{
  \footnotesize{Random generated images of SNGAN and $\LASW$ from  CelebAHQ.
}
} 
  \label{fig:lasw2}
  \vskip -0.1in
\end{figure*}

\begin{figure*}[!t]
\begin{center}
    
  \begin{tabular}{cc}
\widgraph{0.45\textwidth}{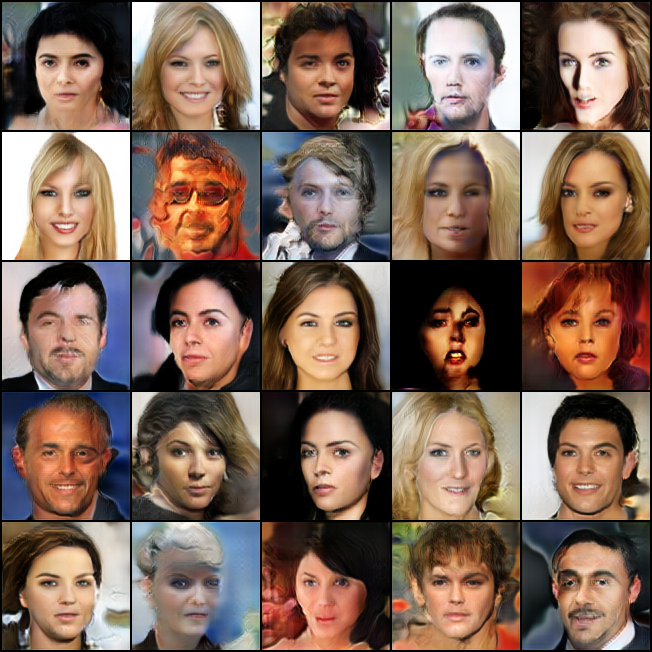} 
&
\widgraph{0.45\textwidth}{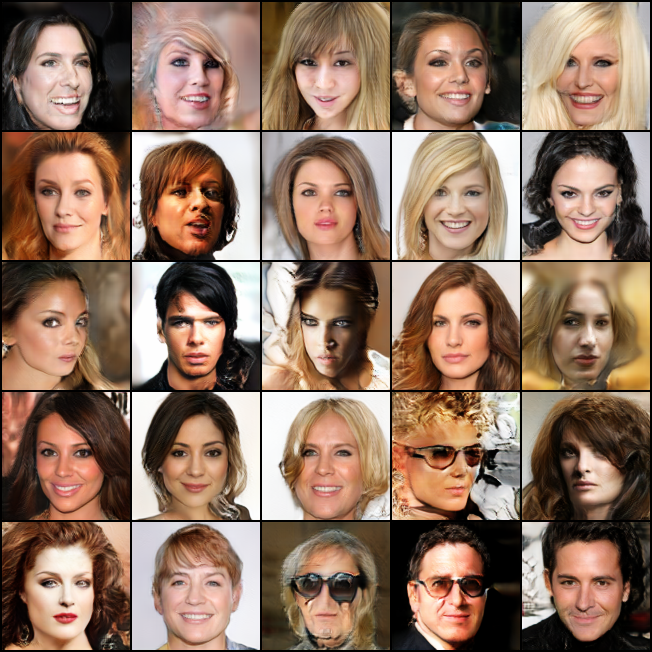} 
\\
Max-SW (CelebA-HQ) & SW (CelebA-HQ)
\\
\widgraph{0.45\textwidth}{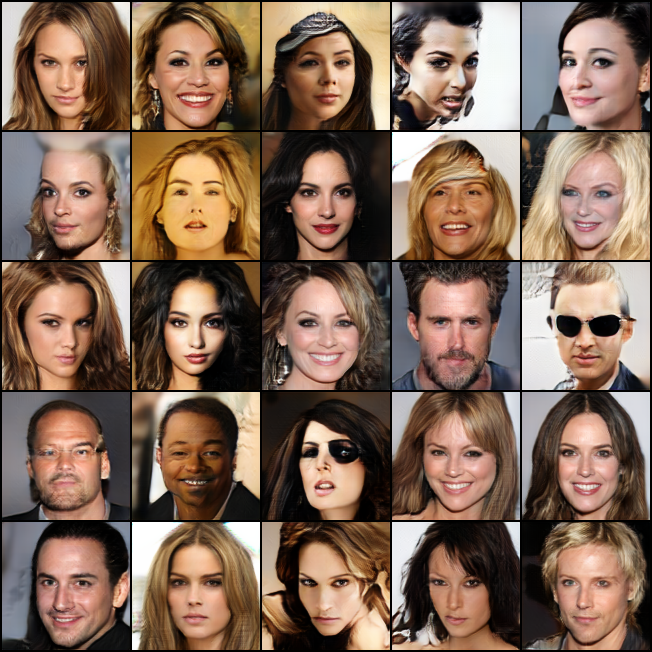} 
&
\widgraph{0.45\textwidth}{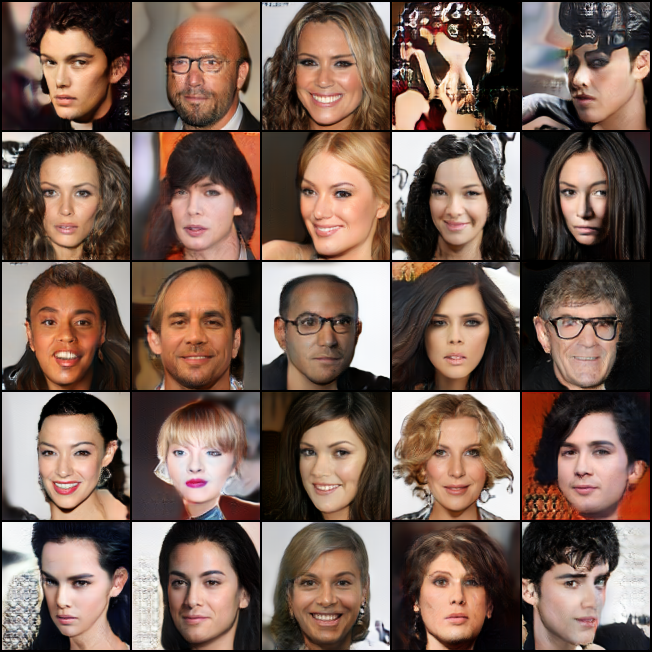} 
\\
$\GASW$ (CelebA-HQ) & $\NASW$ (CelebA-HQ)
  \end{tabular}
  \end{center}
  \vskip -0.1in
  \caption{
  \footnotesize{Random generated images of Max-SW, SW, $\GASW$, and $\NASW$ from  CelebA-HQ.
}
} 
  \label{fig:allhq}
  \vskip -0.1in
\end{figure*}

\textbf{Generated  Images:} We show generated images from SW, $\GASW$, and $\NASW$ on CIFAR10, CelebA, and STL10 in Figure~\ref{fig:sngan1}. The generated images from Max-SW on CIFAR10, CelebA, and STL10  are given in Figure~\ref{fig:sw1}. The generated images from SNGAN and $\LASW$ are given in Figure~\ref{fig:lasw2}. The generated images from SW, Max-SW, $\GASW$, and $\NASW$ on CelebA-HQ are presented in Figure~\ref{fig:allhq}.  Again, we observe consistent quality results compared to the quantitative results of FID scores and Inception scores.

\textbf{Results on Amortized PRW:} We present the result of training generative models on CIfAR10 with mini-batch PRW loss and amortized PRW losses in Table~\ref{tab:detailFIDPRW}. For both PRW and $\mathcal{A}$-PRW, we set the learning rate for $U$ is 0.01. We choose the best result from PRW with the number of gradient updates in $\{10,100\}$ while we only update the amortized model once for $\mathcal{A}$-PRW. We observe that $\mathcal{A}$-PRW gives better FID and IS than PRW for all choice of $k \in \{2,4,16\}$. Moreover, linear amortized projected model gives the best result among amortized models. When $k=16$, the non-linear amortized model suffers from numerical error when using QR decomposition, hence, we cannot provide the result for it. Overall, the result on PRW strengthen the claim that using amortized optimization for deep generative models with (sliced) projected Wasserstein can improve the result.
\begin{table}[]

    \centering
    \caption{\footnotesize{Summary of FID and IS scores of methods based on projected robust Wasserstein on CIFAR10 (32x32).}}
    \scalebox{0.9}{
    \begin{tabular}{l|cc|}
    \toprule
     \multirow{2}{*}{Method}& \multicolumn{2}{c|}{CIFAR10 (32x32)}\\
     & FID ($\downarrow$) &IS ($\uparrow$)\\
           \midrule
           $\text{PRW}$ (k=2) & 42.03 & 6.48\\
         $\mathcal{LA}\text{-PRW}$ (k=2) (ours) & \textbf{14.27} & 8.02\\
           $\mathcal{GA}\text{-PRW}$ (k=2) (ours)  & 14.56 &8.15 \\
          $\mathcal{NA}\text{-PRW}$ (k=2) (ours) & 14.69 &\textbf{8.43}\\
          \midrule
          $\text{PRW}$ (k=4) & 36.82 & 6.50\\
         $\mathcal{LA}\text{-PRW}$ (k=4) (ours) & 14.33& 8.01\\
           $\mathcal{GA}\text{-PRW}$ (k=4) (ours)  & \textbf{13.84} &\textbf{8.18 }\\
          $\mathcal{NA}\text{-PRW}$ (k=4) (ours) & 14.68 &8.05\\
          \midrule
          $\text{PRW}$ (k=16) & 56.74& 5.41\\
         $\mathcal{LA}\text{-PRW}$ (k=16) (ours) & \textbf{14.16}& \textbf{8.06}\\
           $\mathcal{GA}\text{-PRW}$ (k=16) (ours)  & 26.57 &7.31 \\
          $\mathcal{NA}\text{-PRW}$ (k=16) (ours) & - & -\\
         \bottomrule
    \end{tabular}
    }
    \label{tab:detailFIDPRW}
\end{table}
\section{Experimental Settings}
\label{sec:settings}

\vspace{0.5em}
 
\textbf{Neural network architectures:} We present the neural network architectures on CIFAR10 in Table~\ref{tab:netCIFAR}, CelebA in Table~\ref{tab:netCelebA}, STL10 in Table~\ref{tab:netSTL}, and CelebA-HQ in Table~\ref{tab:netLSUN}. In summary, we use directly the architectures  from \url{https://github.com/GongXinyuu/sngan.pytorch}.

\vspace{0.5em}
 
\textbf{Hyper-parameters: } For CIFAR10, CelebA, and CelebA-HQ, we set the training iterations to 50000 while we set it to 100000 in STL10. We update $T_{\beta_1}$ and $T_{\beta_2}$ every iterations  while we update $G_\phi$ each 5 iterations. The mini-batch size $m$ is set to $128$ on CIFAR10 and CelebA, is set to $32$ on STL10, is set to $16$ on CelebA-HQ. The learning rate of $G_\phi$, $T_{\beta_1}$, and $T_{\beta_2}$ is set to $0.0002$. The optimizers for all optimization problems are Adam~\cite{kingma2014adam} with $(\beta_1,\beta_2)=(0,0.9)$.

\vspace{0.5em}
 
\textbf{FID scores and Inception scores:} For these two scores, we calculate them based on 50000 random samples from trained models. For FID scores, the statistics of datasets are calculated on all training samples. 
\begin{table}[!t]
\caption{CIFAR10 architectures.}
\begin{tabular}{ccc}
    \begin{minipage}{.33\linewidth}
        \begin{tabular}{c}
        \toprule
            (a) $G_\phi $\\
            \midrule
            Input: $\boldsymbol{\epsilon} \in \mathbb{R}^{128} \sim \mathcal{N}(0,1)$\\
            \midrule
            $128 \rightarrow 4 \times 4 \times 256 $, dense\\
            linear \\
            \midrule
            $\text { ResBlock up } 256$ \\
            \midrule
            $\text { ResBlock up } 256$ \\
            \midrule
            $\text { ResBlock up } 256$ \\
            \midrule
            $\text { BN, ReLU, } $\\
            $3 \times 3 \text { conv, } 3 \text { Tanh }$\\
            \bottomrule
        \end{tabular}
    \end{minipage} 
    &
    \begin{minipage}{.33\linewidth}
        \begin{tabular}{c}
        \toprule
           (b)  $T_{\beta_1}$ \\
           \midrule
           Input: $\boldsymbol{x} \in[-1,1]^{32 \times 32 \times 3}$ \\
           \midrule
           $\text { ResBlock down } 128$ \\
           \midrule
           $\text { ResBlock down } 128$ \\
           \midrule
           $\text { ResBlock down } 128$ \\
           \midrule
           $\text { ResBlock } 128$ \\
           \midrule
           $\text { ResBlock } 128$ \\
           \bottomrule
        \end{tabular}
    \end{minipage} 
    &
    \begin{minipage}{.33\linewidth}
        \begin{tabular}{c}
        \toprule
           (c)   $T_{\beta_2}$ \\
           \midrule
           Input: $\boldsymbol{x} \in \mathbb{R}^{128 \times 8 \times 8}$ \\
           \midrule
           ReLU \\
           \midrule
           Global sum pooling\\
           \midrule
           $128 \rightarrow 1$\\
           Spectral normalization \\
           \bottomrule
        \end{tabular}
    \end{minipage} 
\end{tabular}
\label{tab:netCIFAR}
\end{table}

\begin{table}[!t]
\caption{CelebA architectures.}
\begin{tabular}{ccc}
    \begin{minipage}{.34\linewidth}
        \begin{tabular}{c}
        \toprule
            (a) $G_\phi $\\
            \midrule
            Input: $\boldsymbol{\epsilon} \in \mathbb{R}^{128} \sim \mathcal{N}(0,1)$\\
            \midrule
            $128 \rightarrow 4 \times 4 \times 256 $, dense\\
            $\text {linear }$ \\
             \midrule
            $\text { ResBlock up } 256$ \\
            \midrule
            $\text { ResBlock up } 256$ \\
            \midrule
            $\text { ResBlock up } 256$ \\
            \midrule
            $\text { ResBlock up } 256$ \\
            \midrule
            $\text { ResBlock up } 256$ \\
            \midrule
            $\text { BN, ReLU, } $\\
            $3 \times 3 \text { conv, } 3 \text { Tanh }$\\
            \bottomrule
        \end{tabular}
    \end{minipage} 
    &
    \begin{minipage}{.33\linewidth}
        \begin{tabular}{c}
        \toprule
           (b)  $T_{\beta_1}$ \\
           \midrule
           Input: $\boldsymbol{x} \in[-1,1]^{64 \times 64 \times 3}$ \\
           \midrule
           $\text { ResBlock down } 128$ \\
           \midrule
           $\text { ResBlock down } 128$ \\
           \midrule
           $\text { ResBlock down } 128$ \\
           \midrule
           $\text { ResBlock } 128$ \\
           \midrule
           $\text { ResBlock } 128$ \\
           \midrule
           $\text { ResBlock } 128$ \\
           \bottomrule
        \end{tabular}
    \end{minipage} 
    &
    \begin{minipage}{.33\linewidth}
        \begin{tabular}{c}
        \toprule
           (c)   $T_{\beta_2}$ \\
           \midrule
           Input: $\boldsymbol{x} \in \mathbb{R}^{128 \times 8 \times 8}$ \\
           \midrule
           ReLU \\
           \midrule
           Global sum pooling\\
           \midrule
           $128 \rightarrow 1$\\
           Spectral normalization \\
           \bottomrule
        \end{tabular}
    \end{minipage} 
\end{tabular}
\label{tab:netCelebA}
\end{table}

\begin{table}[!t]
\caption{STL10 archtectures.}
\begin{tabular}{ccc}
    \begin{minipage}{.34\linewidth}
        \begin{tabular}{c}
        \toprule
            (a) $G_\phi $\\
            \midrule
            Input: $\boldsymbol{\epsilon} \in \mathbb{R}^{128} \sim \mathcal{N}(0,1)$\\
            \midrule
            $128 \rightarrow 3 \times 3 \times 256 $, dense\\
            $\text {, linear }$ \\
            \midrule
            $\text { ResBlock up } 256$ \\
            \midrule
            $\text { ResBlock up } 256$ \\
            \midrule
            $\text { ResBlock up } 256$ \\
            \midrule
            $\text { ResBlock up } 256$ \\
            \midrule
            $\text { ResBlock up } 256$ \\
            \midrule
            $\text { BN, ReLU, } $\\
            $3 \times 3 \text { conv, } 3 \text { Tanh }$\\
            \bottomrule
        \end{tabular}
    \end{minipage} 
    &
    \begin{minipage}{.33\linewidth}
        \begin{tabular}{c}
        \toprule
           (b)  $T_{\beta_1}$ \\
           \midrule
           Input: $\boldsymbol{x} \in[-1,1]^{96 \times 96 \times 3}$ \\
           \midrule
           $\text { ResBlock down } 128$ \\
           \midrule
           $\text { ResBlock down } 128$ \\
           \midrule
           $\text { ResBlock down } 128$ \\
           \midrule
           $\text { ResBlock down } 128$ \\
           \midrule
           $\text { ResBlock } 128$ \\
           \midrule
           $\text { ResBlock } 128$ \\
           \midrule
           $\text { ResBlock } 128$ \\
           \bottomrule
        \end{tabular}
    \end{minipage} 
    &
    \begin{minipage}{.33\linewidth}
        \begin{tabular}{c}
        \toprule
           (c)   $T_{\beta_2}$ \\
           \midrule
           Input: $\boldsymbol{x} \in \mathbb{R}^{128 \times 6 \times 6}$ \\
           \midrule
           ReLU \\
           \midrule
           Global sum pooling\\
           \midrule
           $128 \rightarrow 1$\\
           Spectral normalization \\
           \bottomrule
        \end{tabular}
    \end{minipage} 
\end{tabular}
\label{tab:netSTL}
\end{table}

\begin{table}[!t]
\caption{CelebA-HQ  archtectures.}
\begin{tabular}{ccc}
    \begin{minipage}{.34\linewidth}
        \begin{tabular}{c}
        \toprule
            (a) $G_\phi $\\
            \midrule
            Input: $\boldsymbol{\epsilon} \in \mathbb{R}^{128} \sim \mathcal{N}(0,1)$\\
            \midrule
            $128 \rightarrow 4 \times 4 \times 256 $, dense\\
            $\text {, linear }$ \\
            \midrule
            $\text { ResBlock up } 256$ \\
            \midrule
            $\text { ResBlock up } 256$ \\
            \midrule
            $\text { ResBlock up } 256$ \\
            \midrule
            $\text { ResBlock up } 256$ \\
            \midrule
            $\text { ResBlock up } 256$ \\
            \midrule
            $\text { BN, ReLU, } $\\
            $3 \times 3 \text { conv, } 3 \text { Tanh }$\\
            \bottomrule
        \end{tabular}
    \end{minipage} 
    &
    \begin{minipage}{.33\linewidth}
        \begin{tabular}{c}
        \toprule
           (b)  $T_{\beta_1}$ \\
           \midrule
           Input: $\boldsymbol{x} \in[-1,1]^{128 \times 128 \times 3}$ \\
           \midrule
           $\text { ResBlock down } 128$ \\
           \midrule
           $\text { ResBlock down } 128$ \\
           \midrule
           $\text { ResBlock down } 128$ \\
           \midrule
           $\text { ResBlock down } 128$ \\
           \midrule
           $\text { ResBlock } 128$ \\
           \midrule
           $\text { ResBlock } 128$ \\
           \midrule
           $\text { ResBlock } 128$ \\
           \bottomrule
        \end{tabular}
    \end{minipage} 
    &
    \begin{minipage}{.33\linewidth}
        \begin{tabular}{c}
        \toprule
           (b)   $T_{\beta_2}$ \\
           \midrule
           Input: $\boldsymbol{x} \in \mathbb{R}^{128 \times 8 \times 8}$ \\
           \midrule
           ReLU \\
           \midrule
           Global sum pooling\\
           \midrule
           $128 \rightarrow 1$\\
           Spectral normalization \\
           \bottomrule
        \end{tabular}
    \end{minipage} 
\end{tabular}
\label{tab:netLSUN}
\end{table}
\section{Potential Impact and Limitations}
\label{sec:potential}
\paragraph{Potential Impact:} This work improves training generative models with sliced Wasserstein by using amortized optimization. Moreover, amortized sliced Wasserstein losses can be applied to various applications such as generative models, domain adaptation, and approximate inference, adversarial attack, and so on. Due to its widely used potential, it can be used as a component in some applications that do not have a good purpose. For example, some examples are creating images of people without permission, attacking machine learning systems, and so on.

\paragraph{Limitations:}  In the paper, we have not been able to investigate the  amortization gaps of the proposed amortized models since the connection of the optima of Max-SW to the supports of two probability measures has not been well-understand yet. Moreover, the design of amortized models requires more engineering to achieve better performance since there is no inductive bias for designing them at the moment. The hardness in designing amortized models is that we need to trade-off between the performance and computational efficiency. We will leave these questions to future work. 
\end{document}